\documentclass{article}

\PassOptionsToPackage{numbers, compress}{natbib}

 \usepackage[eandd, final]{neurips_2026}

\usepackage[utf8]{inputenc}
\usepackage[T1]{fontenc}
\usepackage{hyperref}
\usepackage{url}
\usepackage{booktabs}
\usepackage{amsfonts}
\usepackage{amsmath}
\usepackage{amssymb}
\usepackage{nicefrac}
\usepackage{microtype}
\usepackage{xcolor}
\usepackage{xspace}
\usepackage{graphicx}
\usepackage{hyperref}
\usepackage[most]{tcolorbox}
\usepackage{wrapfig}
\usepackage{caption}
\usepackage{enumitem}

\newcommand{\bench}{EmbodiedWorld-200K\xspace}

\title{EmbodiedWorld-200K: Toward Embodied World Agents via Embodied-Planning Dataset and \\ Interactive World Models}

\author{
\textbf{Xiaokun Feng}$^{1,2,4,*}$\hspace{9pt}
\textbf{Junshu Tang}$^{4,*}$\hspace{9pt} 
\textbf{Zeyi Lin}$^{4}$ \hspace{9pt}
\textbf{Ling-Hao Chen}$^{3,4}$\hspace{9pt} \\
\textbf{Shuai Shao}$^{4}$ \hspace{9pt}
\textbf{Qinglin Lu}$^{4}$ \hspace{9pt}
\textbf{Kaiqi Huang}$^{1,2}$\textsuperscript{$\dagger$} \hspace{9pt}
\\
$^1$ School of Artificial Intelligence, UCAS 
$^2$ CASIA 
$^3$ THU
$^4$ Tencent Hunyuan \\
\textbf{Project Page:} \url{https://xiaokunfeng.github.io/EmbodiedWorld-200K/}
}

\begin{document}

\def\thefootnote{}\footnotetext{
Work done during the internship at Tencent Hunyuan. $\quad$ $*$ Equal contribution $\quad$ $\dagger$ Corresponding author
}

\maketitle

\vspace{-2.5em}
\begin{figure}[!h]
    \centering
    \includegraphics[width=0.9\linewidth]{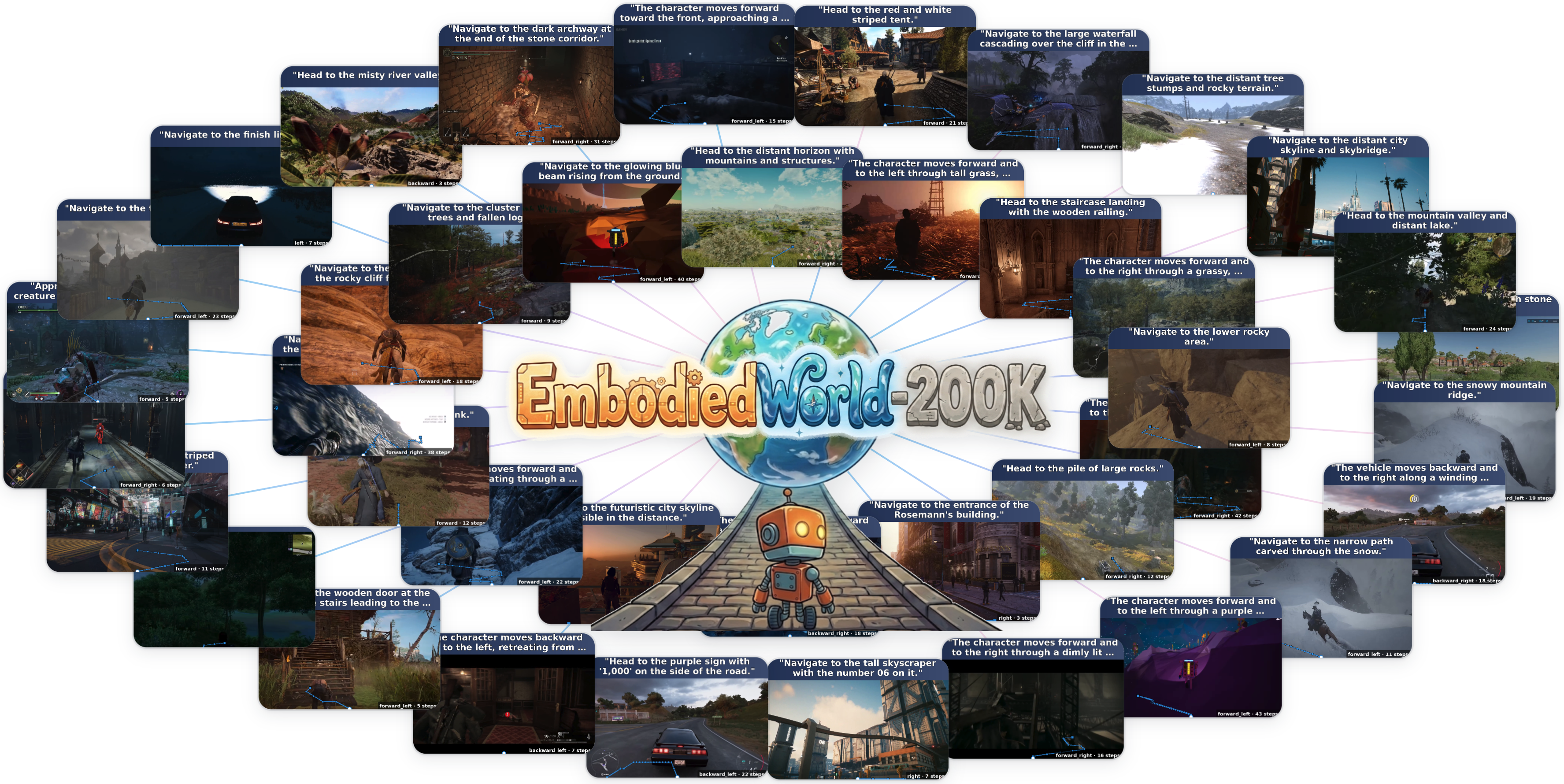}
    \caption{\textbf{Overview of \bench.}
  A large-scale embodied-planning dataset designed to support the study
  of \emph{Embodied World Agents} (EWAs)---general-purpose embodied
  agents operating across diverse open-world 3D environments.
  Each sample consists of an initial visual observation, a
  natural-language navigation instruction, and the corresponding action
  sequence. 
  }
    \label{fig:teaser}
\end{figure}

\begin{abstract}
Building embodied agents that can perceive, reason, and act
across diverse 3D worlds is a central research direction toward
artificial general intelligence. 
Yet most existing work on embodied agents remains confined to specific tasks in a single scene and transfers poorly to new environments, leaving general embodied agents for open-world settings an important but under-explored direction.
We refer to such agents as \textbf{Embodied World Agents (EWAs)}, whose defining property is the ability to understand diverse visual environments and execute general actions in open-world conditions.
To advance research on EWAs, we instantiate them on open-world embodied navigation-planning and release EmbodiedWorld-200K, a large-scale dataset of over 200K samples spanning diverse scenes.
Each example is equipped with multi-level annotations, ranging from low-level camera-motion trajectories to high-level semantic instructions, and is accompanied by unified evaluation splits and task-specific metrics, providing a shared substrate for offline pretraining and fair evaluation of EWAs.
Beyond offline supervision, a defining characteristic of an EWA is its interaction with the environment; training solely on offline data is therefore insufficient, especially for generalizing to scenes not covered in the training distribution. 
Building dedicated real interactive environments, however, is often prohibitively expensive. To address this, we propose a novel training scheme that employs a video world model as a low-cost environment proxy and performs interactive reinforcement-learning (RL) post-training of the EWA within it.
Empirically, pretraining on EmbodiedWorld-200K substantially improves the open-world embodied planning ability of the EWA, and further RL post-training inside a world model yields clear gains on out-of-distribution scenes. 
The full dataset, baseline checkpoints, and the accompanying training and evaluation code for supporting future research on EWAs are publicly available at \url{https://xiaokunfeng.github.io/EmbodiedWorld-200K/}.
\end{abstract}

\vspace{-1em}

\section{Introduction}
\label{sec:intro}

Vision--language large models (VLLMs)
have demonstrated strong semantic understanding and multimodal
reasoning capabilities~\citep{hurst2024gpt,anthropic2024claude,comanici2025gemini,bai2025qwen3,wang2025internvl3}.
However, VLLMs are trained almost exclusively on static, passively
observed data, rather than through executing actions or reasoning about
their consequences in real environments; the intelligence they acquire
thus remains, at its core, disembodied and passive~\citep{silver2025era,sima2},
and falls noticeably short on embodied tasks that require a closed
perception--action loop~\citep{majumdar2024openeqa,fung2025embodied}.
In response, building embodied agents that can perceive, reason,
and autonomously act in diverse 3D
worlds~\citep{kim2024openvla,black2024pi0,rt2} has attracted growing
attention, and is widely regarded as one of the key paths toward
artificial general intelligence.

Yet most existing work on embodied agents remains confined to specific
tasks in a single scene---for example, navigation within a particular
indoor simulator~\citep{anderson2018vision,krantz2020beyond}, manipulation on a particular robotic platform~\citep{openxembodiment2023,kim2024openvla,black2024pi0,intelligence2504pi0}, or task
execution within a single type of game~\citep{fan2022minedojo,baker2022video,lifshitz2023steve,bytedance2025lumine,lee2022multi}. By contrast, developing
general-purpose embodied agents for broad open-world environments
remains an important but under-explored direction. For ease of
discussion, we refer to such agents as \textbf{Embodied World Agents
(EWAs)}: agents that aim to acquire both a general understanding of
diverse visual environments in open worlds and a general ability to
execute actions within them. While this vision is highly appealing,
realizing it faces a fundamental infrastructure gap: the community
currently lacks a large-scale, open-world dataset that can
simultaneously support well-grounded training and unified
evaluation, and a matching open-source baseline model together
with a practical training recipe.


Existing embodied-agent datasets, such as typical Vision-and-Language Navigation
(VLN) benchmarks~\citep{anderson2018evaluation,krantz2020beyond}, are limited in scale, lack scene diversity, and offer only single-level annotations, preventing EWAs from learning robust, general-purpose priors through large-scale pretraining as VLLMs do.
To fill this gap, we instantiate the EWA problem as \textbf{open-world
embodied navigation-planning}, a task that jointly stresses perceiving a
diverse visual scene, grounding a language goal within it, and executing
an extended action sequence. 
We adopt this instantiation because
translation and view rotation are the actions that remain universal
across heterogeneous open worlds, and are also the action types that
frontier video world models natively support~\citep{bruce2024genie,gamecraft1,team2026advancing}. 
Accordingly, we release
\textbf{EmbodiedWorld-200K}, a large-scale public corpus for this task,
comprising over 200K video clips collected from diverse open-world
virtual game environments (as shown in Fig.~\ref{fig:teaser}).
Each sample is accompanied by multi-level annotations, ranging
from low-level camera-motion/action trajectories to high-level semantic
instructions that describe the underlying embodied navigation-planning task.
Building on this large-scale corpus of instruction--action sequences, we pretrain an existing VLLM~\citep{bai2025qwen3} in an offline manner, retaining its multimodal understanding while extending it into a Vision–Language–Action (VLA) model \citep{sapkota2025vision,kim2024openvla} with open-world embodied navigation-planning capabilities.

A defining property of EWAs is their interaction with open-world environments, where many required capabilities can only be acquired through closed-loop interaction.
Consequently, training solely on EmbodiedWorld-200K in an offline manner is insufficient, particularly for generalization to unseen scenes. However, real-world interactive platforms, such as game engines, are limited in scalability across diverse environments.
Fortunately, the rapid progress of video-generative world models (WMs)~\citep{genie3,cosmos2025,matrixgame2025,pan2025,gamecraft1,gamecraft2,hyworld2025,team2026advancing}  opens a new avenue: modern WMs can already roll out future visual  observations conditioned on actions, offering a flexible and scalable simulation substrate for open-world interaction across diverse scenes.
Building on this capability, we propose a novel training paradigm that treats a video WM as an environment proxy and performs interactive reinforcement learning (RL) post-training of EWAs within it. Specifically, an offline-pretrained EWA generates trajectories inside a frontier WM. During rollout, a VLLM dynamically produces task instructions conditioned on the evolving environment semantics, while reward signals are derived by framing reward modeling as visual question answering (QA)~\citep{rocamonde2024vlmrm,wang2024rlvlmf}. Finally, we apply direct preference optimization (DPO)~\citep{rafailov2023dpo} on the collected trajectories, enabling the model to refine its policy from interactive feedback and improving embodied planning in unseen environments.

Extensive experiments demonstrate that offline pretraining on EmbodiedWorld-200K significantly improves the open-world embodied planning ability of EWAs, and that subsequent interactive RL post-training within a video world model further enhances adaptation to out-of-distribution (OOD) scenes. Our main contributions are summarized as follows:
\begin{itemize}[leftmargin=1em]
  \item We study general-purpose embodied agents for open-world environments, referred to as Embodied World Agents (EWAs), and position them as an important yet under-explored research direction. We further provide infrastructure support from both \textbf{data} and \textbf{training} perspectives.

  \item We release \textbf{EmbodiedWorld-200K}, a large-scale dataset for open-world embodied planning containing over 200K samples with multi-level annotations, together with a unified evaluation split and task-specific metrics, enabling standardized pretraining and fair evaluation of EWAs.

  \item We propose a \textbf{two-stage training framework} for EWAs and instantiate a corresponding baseline: offline pretraining on EmbodiedWorld-200K followed by interactive RL post-training using a video world model as a scalable environment proxy.

  \item Empirically, EmbodiedWorld-200K significantly improves embodied planning performance, while WM-based interactive post-training further enhances generalization to unseen scenes. We release the dataset, model weights, and code at the \href{https://xiaokunfeng.github.io/EmbodiedWorld-200K/}{\textcolor{blue}{project page}} to support future research.
\end{itemize}

 \section{Related Work}                                 
  \label{sec:related}   
                     
  \paragraph{Embodied agents.}
  Building agents that perceive and act in 3D environments has been
  studied across multiple domains. In \emph{robotic manipulation}, 
  RT-2~\citep{rt2} finetunes a vision–language model to directly
  output robotic actions, PaLM-E~\citep{driess2023palme} ingests  
  multi-modal sensor streams into a large language model for embodied
  planning, SayCan~\citep{saycan2022} grounds language plans through 
  learned affordances, and OpenVLA~\citep{kim2024openvla} and
  $\pi_0$~\citep{black2024pi0} scale open-source vision–language–action
 (VLA) policies with large robot datasets
  such as Open X-Embodiment~\citep{openxembodiment2023}. In      
  \emph{embodied navigation}, the community has long relied on
  photo-realistic simulators with curated
  trajectories~\citep{anderson2018vision,krantz2020beyond}, yet these
  benchmarks remain limited to a few thousand paths in indoor scans. In
  \emph{open-world virtual environments}, VPT~\citep{baker2022video}
  learns policies from unlabeled gameplay videos,
  MineDojo~\citep{fan2022minedojo} provides Internet-scale knowledge
  for open-ended agents, Voyager~\citep{wang2023voyager} demonstrates
  LLM-driven lifelong exploration, STEVE-1~\citep{lifshitz2023steve}
  trains a text-conditioned policy via latent goals, and
  SIMA~\citep{sima1} shows instruction-following across multiple 3D
  games. Despite this progress, most efforts remain confined to a
  single task type, scene distribution, or platform.  The prospect of
  \emph{general-purpose} embodied agents operating across diverse open
  worlds---what we term Embodied World Agents (EWAs)---remains largely
  unexplored, primarily due to the absence of large-scale, multi-level
  annotated data and a matching training recipe.

  \paragraph{World (action) models.}
  World models learn environment dynamics to support agent training or
  planning. Classic model-based RL methods such as
  DreamerV3~\citep{hafner2023dreamerv3} learn compact latent dynamics
  for policy optimization, while IRIS~\citep{micheli2023iris} builds
  autoregressive world models with transformers. Recent
  \emph{video-generative} world models move toward pixel-level
  fidelity: Genie~\citep{bruce2024genie} learns controllable
  interactive environments from unlabeled video,
  GameNGen~\citep{valevski2024gamengen} and
  DIAMOND~\citep{alonso2024diamond} simulate game engines with
  diffusion models, GAIA-1~\citep{hu2023gaia1} generates driving
  scenarios, and Cosmos~\citep{cosmos2025} provides a large-scale
  foundation world model for physical AI. UniSim~\citep{yang2023learning}
  further proposes learning interactive real-world simulators that unify
  visual, physical, and semantic signals.
  Connecting world models to downstream agents has been explored through
  several paradigms. \emph{Planning-based} approaches use video
  generation as an imagination module for lookahead: Video Language
  Planning~\citep{du2023videolanguageplanning} leverages text-to-video
  models to synthesize visual plans, and
  RoboDreamer~\citep{zhou2024robodreamer} composes world models for
  robot imagination.  \emph{Data-augmentation} methods generate
  synthetic trajectories for offline policy improvement, as in
  UniPi~\citep{du2023unipi}, which casts policy learning as
  text-guided video generation, and
  GenRL~\citep{mazzaglia2024genrl}, which trains embodied policies
  inside a multimodal foundation world model. The paradigm closest to
  ours treats the world model as a \emph{closed-loop interactive
  environment for RL}: UniSim~\citep{yang2023learning} trains reward and
  policy jointly within a learned simulator, and
  GR-2~\citep{cheang2024gr2} uses generated video for a scalable robot
  post-training. However, these works predominantly target
  \emph{robotic manipulation} with low-level motor commands, or operate
  on a single platform. Our work differs in both task scope and
  contribution form: we target embodied navigation and planning across
  general open-world scenes, and rather than proposing a new world
  model or RL algorithm, we provide reusable research infrastructure, a
  large-scale corpus (\bench{}), a concrete task formulation, and a
  reproducible two-stage training recipe (offline pretraining + WM-based
  RL post-training). These efforts provide precisely the shared foundation left open
  by prior work.

\section{The \bench Dataset}
\label{sec:dataset}



We introduce \bench, a large-scale dataset for pretraining \emph{general-purpose embodied world agents} across diverse open-world environments. Given that virtual 3D games have long served as a standard testbed for embodied-agent research~\citep{kim2024openvla,black2024pi0,rt2}, \bench is assembled from 150 3D open-world games spanning a wide range of genres, yielding over 200K gameplay trajectories organized under a unified interaction format.
Building on this corpus, we adopt open-world embodied navigation-planning as the central task for studying Embodied World Agents (EWAs), as it effectively stresses the ability of an agent to perceive, reason over, and act within diverse 3D environments. Targeted at this task, we apply a structured annotation process to the raw gameplay data. Specifically, each trajectory in \bench is converted into a triplet consisting of an initial visual observation, a paired natural-language task instruction that specifies a diverse navigation-planning goal, and the expected action sequence the agent is supposed to execute in response.
We detail the construction of EmbodiedWorld-200K in the following sections.

\begin{figure}[!t]
    \centering
     \vspace{-3em}
    \includegraphics[width=\linewidth]{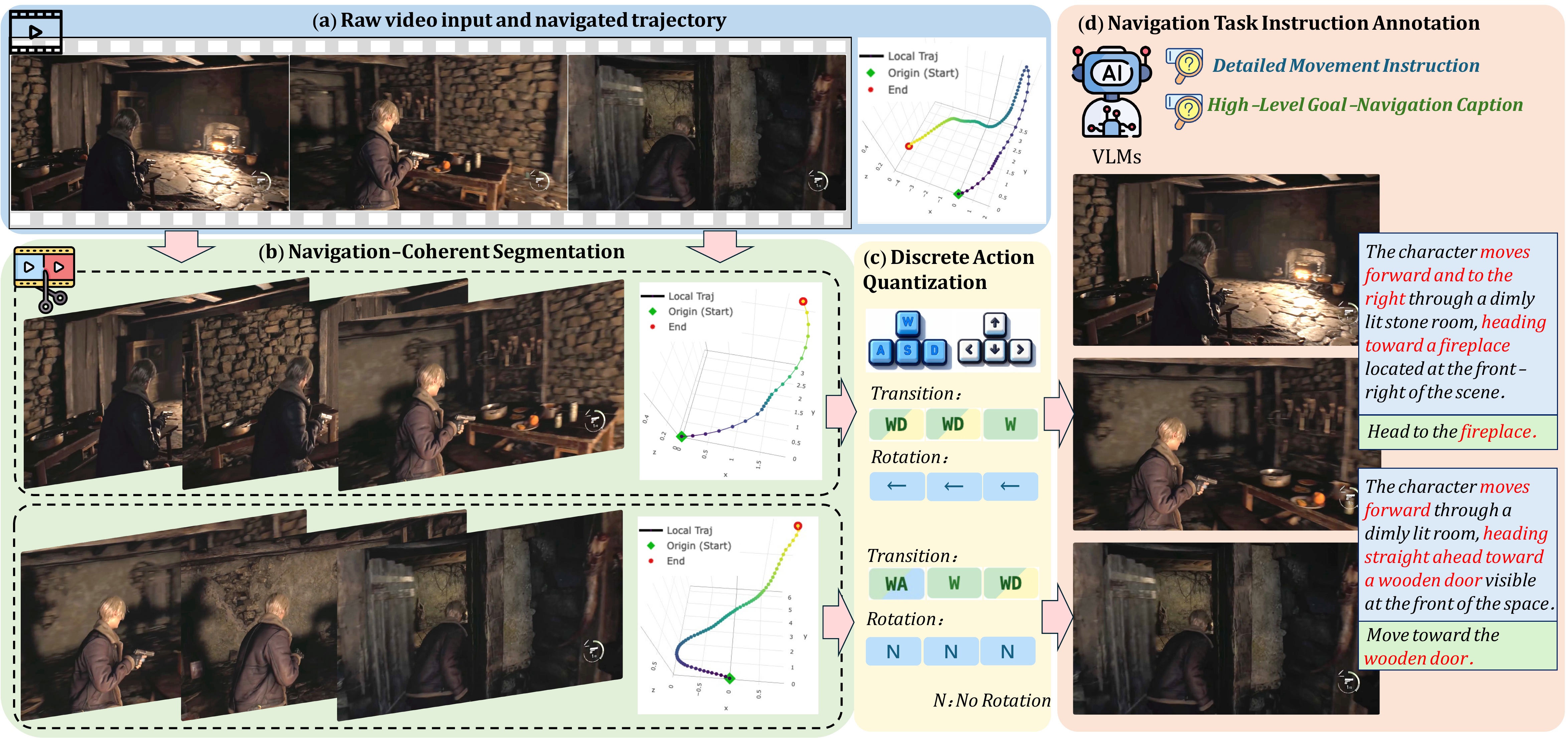}
    \caption{\textbf{Data annotation pipeline.} Given \textbf{(a)} the raw video
  and its camera-motion trajectory, we first \textbf{(b)} leverage the trajectory
  to perform navigation-coherent segmentation; then on each resulting
  segment we carry out \textbf{(c)} discrete action quantization and \textbf{(d)} navigation task
  instruction annotation.}
    \label{fig:data_pipeline}
\vspace{-1em}
\end{figure}

\subsection{Data Collection}



To support research on EWAs, the training corpus should cover a wide variety of 3D environments and enable the agent to perform a rich set of interactive tasks within them. However, few existing embodiment-related datasets meet both requirements simultaneously. Especially, datasets for robotic manipulation~\citep{kalashnikov2018scalable,rt2,o2024open,walke2023bridgedata,team2024octo} focus on controlling a specific robot platform within relatively uniform environments, establishing VLN benchmarks~\citep{anderson2018evaluation,krantz2020beyond} are largely restricted to indoor scenes. Besides, autonomous-driving corpora~\citep{cosmos2025,nvidia2025cosmostransfer1conditionalworldgeneration} are confined to the special case of road traffic. None of these can therefore be reused directly for our research. In contrast, virtual 3D games have long served as a natural proving ground for embodied-agent research~\citep{wang2023voyager,wang2024jarvis,tan2024cradle,bytedance2025lumine,sima1,sima2}. Modern 3D open-world games offer visually rich environments that obey physically plausible dynamics, and typically adopt first- or third-person viewpoints that align naturally with those of an embodied observer. More importantly, under a unified interaction logic, they jointly cover a broad spectrum of embodied behaviors such as navigation, exploration, and object interaction~\citep{cheng2025embodiedeval,zhu2025understand3dscenebridging}, which makes them a highly suitable data platform for studying EWAs.

Motivated by this observation, we choose to build EmbodiedWorld-200K from gameplay data. Specifically, we adopt OGameData~\citep{che2024gamegen} as our data foundation: a publicly available, large-scale game-video repository that spans more than 150 next-generation 3D games and provides YouTube video URLs, temporal annotations, and structured captions for over one million gameplay clips, from which a multi-stage construction pipeline yields high-quality clips that are largely UI-free and strongly interaction-relevant. However, since OGameData was originally curated for open-world game video generation rather than for embodied-agent research, we further adapt it to the EWA setting through an additional filtering step. Concretely, we retain only clips that exhibit clear and continuous embodied actions, while discarding passive content such as cutscenes and menu navigation. To preserve genre-level diversity while staying aligned with the EWA task, we concentrate on open-world exploration, action-adventure, first-person shooter (FPS), puzzle-solving, and simulation games. After this process, we obtain 260K gameplay trajectories ($\sim$708 hours in total), which serve as the raw input to the annotation pipeline described in the next subsection.

\subsection{Data Processing and Annotation Pipeline}

The core contribution of \bench lies in converting raw gameplay videos into structured agent training data via a dedicated annotation pipeline, illustrated in Fig.~\ref{fig:data_pipeline}.

\paragraph{Task Formulation.}
We instantiate the EWA study through the \textbf{open-world embodied navigation-planning} task, because it simultaneously stresses the three core abilities a general EWA must possess---perceiving a diverse visual environment, reasoning over the relation between goal and scene, and executing an extended action sequence within the world. Concretely, given a visual context in a 3D scene together with a natural-language goal instruction, the agent is required to produce the sequence of actions that fulfills this goal. Formally, an EWA carries out this task as
\[
\mathbf{a}_{1:T} \;=\; \mathrm{Agent}\!\left(o_0,\, \ell\right),
\]
where $o_0$ denotes the initial visual observation of episode, $\ell$ is the natural-language task instruction describing a navigation-planning goal, and $\mathbf{a}_{1:T} = (a_1, a_2, \dots, a_T)$ is the expected action sequence to execute.
Accordingly, each basic sample in the EWA task takes the form of a triplet $(o_0,\, \ell,\, \mathbf{a}_{1:T})$, and our annotation pipeline is organized around the construction of these three components.

As a preprocessing step, we first recover camera motion from each video clip using VIPE~\citep{huang2025vipe}, to obtain 6-DoF pose trajectory capturing both translation and rotation. Together with raw gameplay videos, these camera trajectories form the input to the subsequent annotation pipeline (Fig.~\ref{fig:data_pipeline} (a)).

\begin{figure}[!t]
    \centering
    \includegraphics[width=\linewidth]{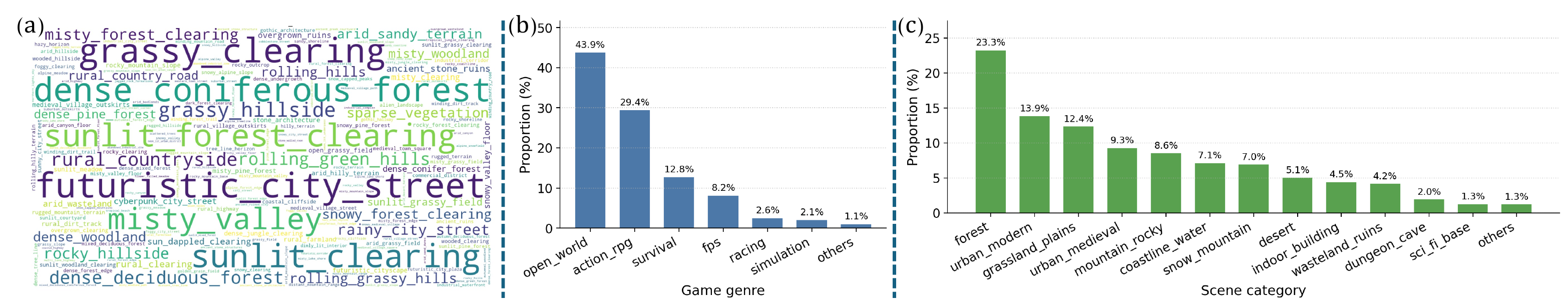}
\vspace{-1.5em}
    \caption{\textbf{Statistical analysis of environment types.}
     \textbf{(a)} Word cloud over the descriptive keywords of the observed environments. \textbf{(b-c)} Distribution of NCSs by game genre and by scene category.}
\vspace{-1em}
    \label{fig:SA_scene_1}
\end{figure}

\paragraph{Step 1: Navigation-Coherent Segmentation and Extraction of $o_0$.}
Although OGameData already provides scene-cut clips, they are not directly usable as EWA navigation units: within a single clip the subject's motion can be visually incoherent, such as mixing drifting, backtracking, and frequent direction change. As a result, there is no single goal-directed trajectory whose intent we can attach an instruction to. Since navigation is naturally defined by a consistent principal motion direction towards some destination, we first segment each OGameData clip into shorter pieces that each carry one dominant motion direction. We refer to each such piece as a \textbf{navigation-coherent segment (NCS)}, and adopt the NCS as the basic analysis unit throughout EmbodiedWorld-200K.

Concretely, starting from the camera pose sequence of a clip, we compute a smoothed per-frame displacement, track the cumulative principal motion direction, and trigger a cut whenever the instantaneous motion deviates from this principal direction by more than a preset angle for several consecutive frames. After further discarding erratic and overly short candidates, we obtain in total 215K NCSs from the raw gameplay clips. Each surviving NCS carries a well-defined principal motion direction, \textit{i.e.}, a clear navigation intent. Therefore, we directly take its initial frame as the visual observation $o_0$ of the corresponding EWA sample. The exact thresholds and full implementation details of this procedure are deferred to App.~\ref{app:ncs}.

\paragraph{Step 2: Discrete Action Quantization and Derivation of $\mathbf{a}_{1:T}$.}

For every NCS, we derive the agent action sequence $\mathbf{a}_{1:T}$ by discretizing the camera-pose motion into a unified W/A/S/D action space shared across all games:
\begin{itemize}[leftmargin=1em]
    \item \textbf{Translation:} W (forward), A (left), S (backward), D (right), N (stationary)
    \item \textbf{Rotation:} W (pitch up), A (yaw left), S (pitch down), D (yaw right), N (no rotation)
\end{itemize}
The central design question is how to calibrate the physical amount of motion that each translation/rotation token should represent. This calibration directly controls the distribution of $T$ across NCSs: too coarse a unit yields sequences that are too short to carry fine-grained navigation signals, whereas too fine a unit produces sequences that are too long to learn efficiently. Therefore, we initially run a dataset-wide statistic on the per-NCS motion magnitude, and pick a translation unit and a rotation unit from this distribution that make the resulting $T$ lie in a sensible range. With these two units fixed, we scan each NCS in temporal order and emit a new W/A/S/D token whenever the accumulated motion reaches one unit, independently for translation and rotation, so that the two streams remain aligned. As a result, every NCS is converted into a variable-length W/A/S/D sequence in which each token corresponds to an approximately fixed amount of physical motion. Detailed statistics, the exact discretization procedure, and additional robustness safeguards are provided in App.~\ref{app:quant}.

\paragraph{Step 3: Annotation of the Navigation Task Instruction $\ell$.}
Given the initial visual observation $o_0$ and the direction-coherent action sequence $\mathbf{a}_{1:T}$, the remaining component of the triplet, the natural-language task instruction $\ell$, is obtained by automatic prompting of a frontier vision-language model, Qwen3.5-27B~\citep{bai2025qwen3}. For each NCS we feed into the VLLM a set of frames uniformly sampled from the corresponding video segment, together with an overall direction descriptor derived from $\mathbf{a}_{1:T}$, and prompt the model to produce a natural-language instruction that summarizes the embodied-navigation task implicitly carried by that segment. To accommodate different annotation granularities, we design two complementary instruction types:

\begin{itemize}[leftmargin=1em]
    \item \textbf{Detailed Movement Instruction.} A fine-grained description of how the agent should move throughout the segment, combining direction cues with the visible target---\textit{e.g.}, ``The character advances forward along a narrow dirt path, heading toward the line of trees visible directly ahead.''

    \item \textbf{High-Level Goal-Navigation Instruction.} A direction-agnostic description that specifies only the destination to reach, with no movement cue attached, \textit{e.g.}, ``Head to the line of trees.'' or ``Navigate to the wooden gate structure.'' Compared with the detailed instruction above, this higher-level variant is noticeably more challenging. Moreover, it is not always producible, because it requires the navigation target to be uniquely identifiable within the segment.  Therefore, we design an automatic grounding--verification procedure admitting only the NCSs satisfying this condition into this type of annotation.
\end{itemize}

\begin{figure}[!t]
    \centering
    \includegraphics[width=\linewidth]{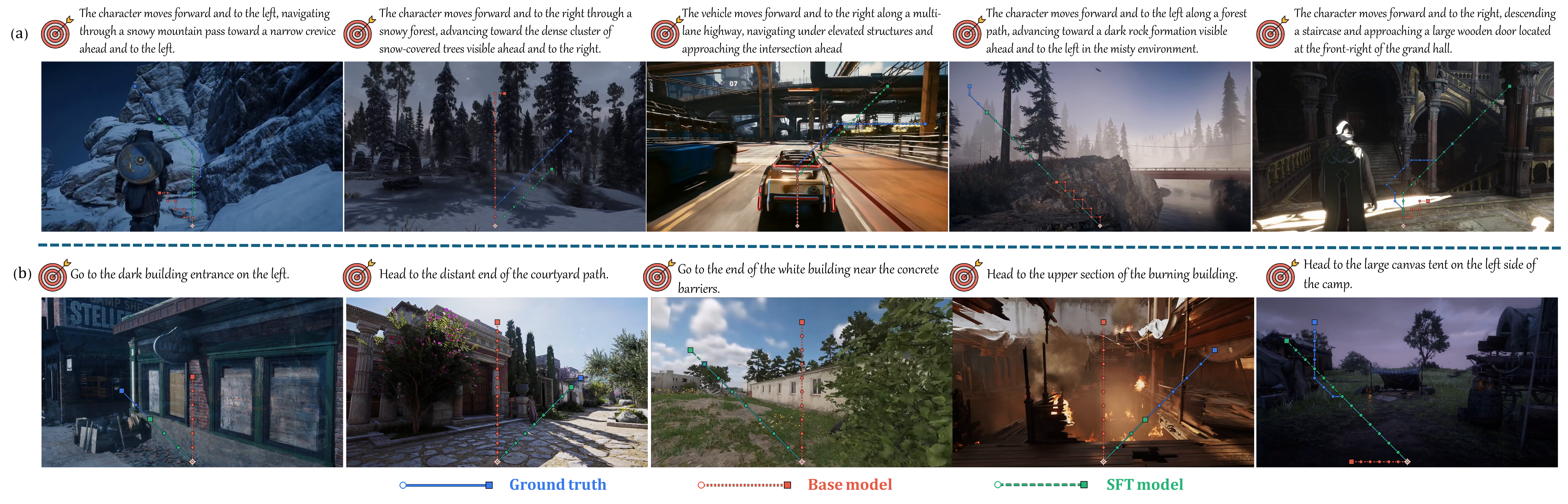}
\vspace{-2em}
\caption{\textbf{Qualitative comparison examples before and after Stage-I pretraining}, under \textbf{(a)} detailed movement instructions and \textbf{(b)} high-level goal-navigation instruction.
  }
\vspace{-1em}
    \label{fig:vis_case_sft}
\end{figure}

When constructing the final training data, for each NCS we adopt a higher-level-first, detailed-fallback policy: we use the goal-navigation caption as $\ell$ whenever it is successfully produced, and otherwise fall back to the detailed movement instruction. 
Fig.~\ref{fig:vis_case_sft} (a) and (b) present representative examples of the Detailed Movement      
Instruction and the High-Level Goal-Navigation Instruction, respectively, and more examples can be found 
in Fig.~\ref{fig:teaser}.
The full prompts, the grounding--verification procedure, and the post-hoc consistency-check rules are deferred to App.~\ref{app:annot}.



\begin{figure}[!t]                                                                                      
  \centering
  \begin{minipage}[t]{0.54\linewidth}                                                                 
      \centering
      \includegraphics[width=0.95\linewidth]{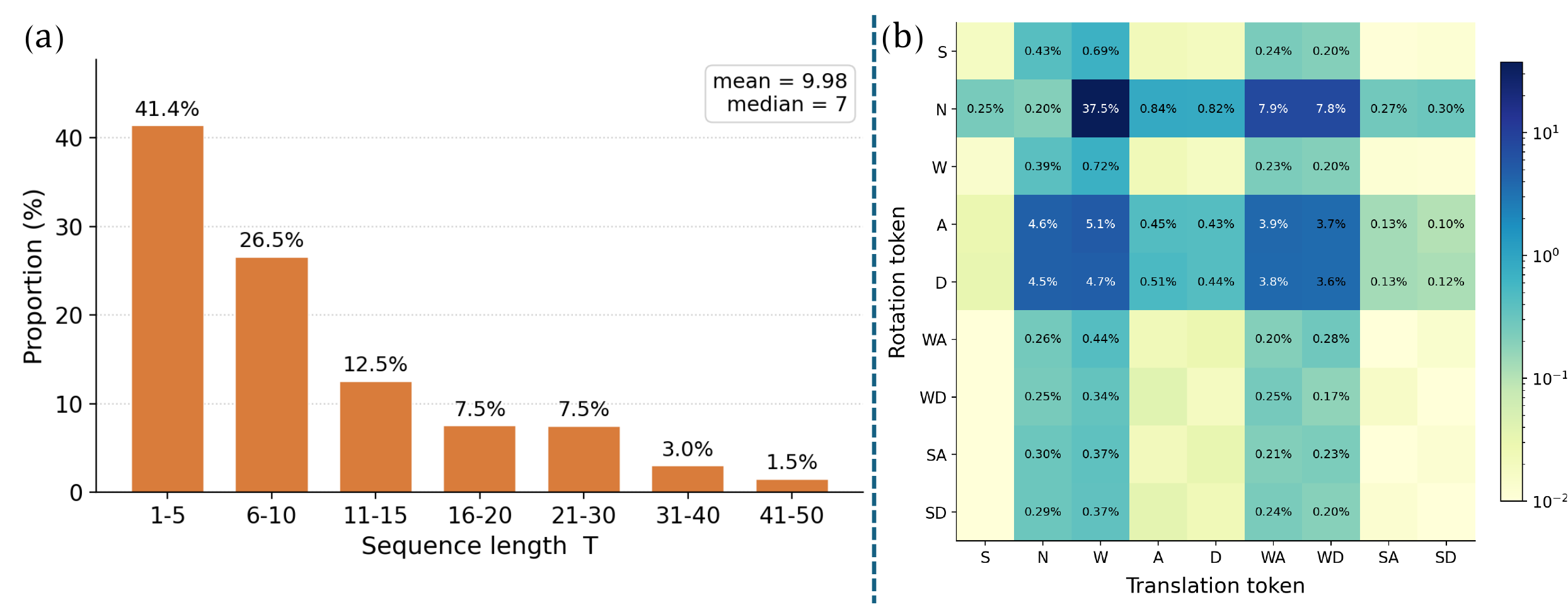}
      \caption{\textbf{Statistical analysis of action sequence $\mathbf{a}_{1:T}$.}
       \textbf{(a)} Distribution of sequence lengths $T$ across all NCSs.
       \textbf{(b)} Per-position heatmap of action-token occurrences.}
      \label{fig:SA_action}
  \end{minipage}
  \hfill
  \begin{minipage}[t]{0.4\linewidth}
      \centering
      \includegraphics[width=\linewidth]{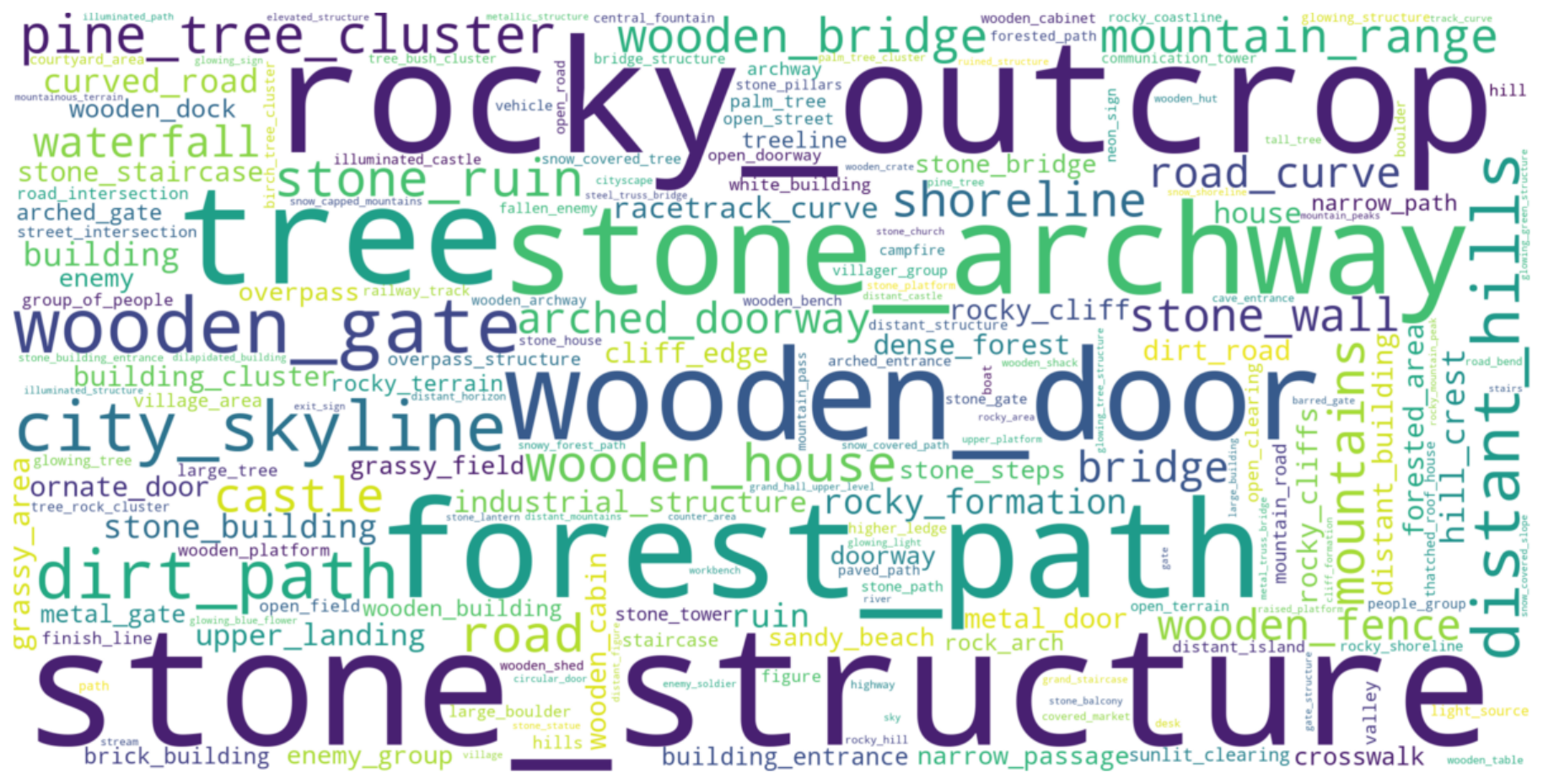}
      \caption{\textbf{Statistical analysis of the navigation goals.}}
      \label{fig:SA_goal}
  \end{minipage}
\vspace{-1em}
\end{figure}

\subsection{Statistical Analysis }
Based on the NCSs extracted above, we carry out a statistical analysis around the core triplet $(o_0,\, \ell,\, \mathbf{a}_{1:T})$, to offer a direct view of how well EmbodiedWorld-200K matches the needs of EWA research.
Firstly, the visual observation $o_0$ reflects the type of world each sample is drawn from, and is thus the most direct lens for evaluating environmental diversity. Fig.~\ref{fig:SA_scene_1} (a) shows a word cloud over the descriptive keywords of the observed environments, and Fig.~\ref{fig:SA_scene_1} (b–c) provide finer-grained breakdowns along the game-genre and scene-category axes. Complementary per-dimension statistics (camera viewpoint, art style, etc.) are reported in  Fig.~\ref{fig:SA_scene_app}. Overall, these plots indicate that EmbodiedWorld-200K spans a wide variety of 3D environments rather than being dominated by any single scene type.

Besides, the action sequence $\mathbf{a}_{1:T}$ captures the granularity and length of the operations the agent is expected to execute. Fig.~\ref{fig:SA_action} (a) plots the distribution of sequence lengths $T$, while Fig.~\ref{fig:SA_action} (b) shows a heatmap of per-position token occurrences across all NCSs. The fraction of samples decreases as $T$ grows, which is consistent with our design choice that the navigation target typically appears in the initial frame $o_0$. That is to say, the agent–target distance therefore stays moderate and does not call for excessively long sequences. The mean length of 9.98 serves as an appropriate operating point, covering both short-range fine adjustments and longer-range continuous progressions.
Finally, for the task instruction $\ell$, we primarily examine whether the set of navigation goals is sufficiently rich. Fig.~\ref{fig:SA_goal} shows a word cloud over the target-location phrases extracted from $\ell$, confirming that the covered goals span a wide variety of object and place categories.
The above analyses collectively demonstrate that EmbodiedWorld-200K provides rich diversity and broad coverage across its three core components (visual environments, action sequences, and task instructions), offering a solid data foundation on which EWA research can be built.

\subsection{Evaluation Metrics}
\label{sec:metric}
We partition the curated NCSs into disjoint training and evaluation splits so that all methods are trained and benchmarked under a unified protocol~\citep{feng2026narrlv}. 
Our metrics are designed to quantitatively assess instruction-conditioned trajectory following and goal arrival~\citep{ling2025vmbench}.
Drawing on standard practice in Vision-and-Language Navigation (VLN) and tailored to our discrete W/A/S/D action format, we report four complementary metrics that span token-level and trajectory-level performance (see App.~\ref{app:metrics} for algorithmic details): 

\begin{itemize}[leftmargin=1em]
    \item \textbf{TM (Token Match, $\uparrow$)}: mean per-step keyset-F1 between predicted and ground-truth tokens, computed separately on the translation and rotation streams and then averaged.
    \item \textbf{DirAcc (Direction Accuracy, $\uparrow$)}: global-direction accuracy averaged over four axes (translation forward/backward and left/right, rotation pitch and yaw).
    \item \textbf{nDTW (normalized Dynamic Time Warping, $\uparrow$)}: normalized DTW over reconstructed 3D trajectories, averaged over a distance-tolerance coefficient $\tau \in \{0.5, 1.0, 2.0\}$ that controls how strictly per-step deviations are penalized.
    \item \textbf{NE (Navigation Error, $\downarrow$)}: terminal xy displacement measured in translation units.
\end{itemize}

\section{EWA Baseline Method}
\label{sec:baselines}

To facilitate future research on EWAs, we present a simple baseline that turns a frontier VLLM into an EWA through a dedicated two-stage training scheme. Concretely, we conduct large-scale offline pretraining on EmbodiedWorld-200K, and then perform interactive RL post-training with a video world model acting as environmental proxy. We describe each component as follows.

\subsection{Model Architecture}
Although frontier VLLMs have demonstrated strong multimodal understanding, the intelligence they acquire remains essentially disembodied: they observe passively and lack a native mechanism for producing physical actions in the environment~\citep{sima2}. Our baseline therefore takes a frontier VLLM (\textit{i.e.}, Qwen-VL~\citep{bai2025qwen3}) as backbone and extends it into a VLA model for open-world embodied planning. We deliberately refrain from introducing heavyweight extra components such as dedicated action heads~\citep{li2024cogact,black2024pi0,hou2024diffusion}, and instead keep both the task instruction and the predicted action sequence entirely within VLLM native text-token space. This minimally intrusive design preserves the understanding capability inherited from VLLM pretraining while providing a simple, easy-to-extend baseline for follow-up work. Concretely, the visual observation $o_0$ and the task instruction $\ell$ are concatenated through a dedicated system prompt as input, and the model is asked to emit the full action sequence $\mathbf{a}_{1:T}$ as a single JSON object. Detailed prompt is provided in App~\ref{app:stage1}.

\subsection{Two-Stage Training}
To equip VLLM with open-world embodied-planning ability, we first perform offline pretraining on our EmbodiedWorld-200K dataset so that the model can absorb the open-world embodied-planning priors latent in this large-scale corpus. Building on this offline-pretrained EWA, we further introduce a second
stage of interactive RL post-training. This stage targets adaptation to
specific environments, especially scenes outside the training
distribution, where closed-loop interaction is crucial. To make such
interaction scalable, we use a video world model~\citep{team2026advancing} as a low-cost
environment proxy.

\paragraph{Stage I: Offline supervised pretraining.} We finetune Qwen3-VL-4B~\citep{bai2025qwen3} on the training split of EmbodiedWorld-200K with a cross-entropy loss between the predicted and ground-truth action tokens. This stage injects broad open-world embodied-planning priors — drawn from a diverse distribution of scenes and instructions — into the model.

\paragraph{Stage II: Interactive RL post-training inside a video world model.} Stage I is purely offline, and its coverage is therefore bounded by the
training distribution. To further adapt the baseline to a particular
target environment, especially one outside the coverage of
EmbodiedWorld-200K, closed-loop interaction is necessary. Since building
real interactive platforms at open-world scale is prohibitively
expensive, we instead leverage a frontier video world
model~\citep{team2026advancing,genie3,matrixgame2025,feng2026enhancing} as a low-cost environment proxy. The
Stage-I baseline is rolled out inside this proxy environment, and the
resulting trajectories are used for RL post-training.
Although conceptually straightforward, realising this pipeline requires careful design at every step. The core recipe of our scheme is as follows: for each rollout, a VLLM is conditioned on the current scene to adaptively generate a natural-language task instruction $\ell$, which in turn anchors a visual question-answering (VQA) reward that judges whether the rollout fulfils the instruction; the resulting preference trajectory pairs are then consumed by Direct Preference Optimization (DPO)~\citep{rafailov2023dpo,chen2026taming} to further post-train the baseline. In this way, we aim to substantially enhance the baseline's embodied-planning ability in specific target scenes without requiring any additional human annotation. Further implementation details are provided in App.~\ref{app:stage2}.

\begin{figure}[!t]
    \centering
    \includegraphics[width=\linewidth]{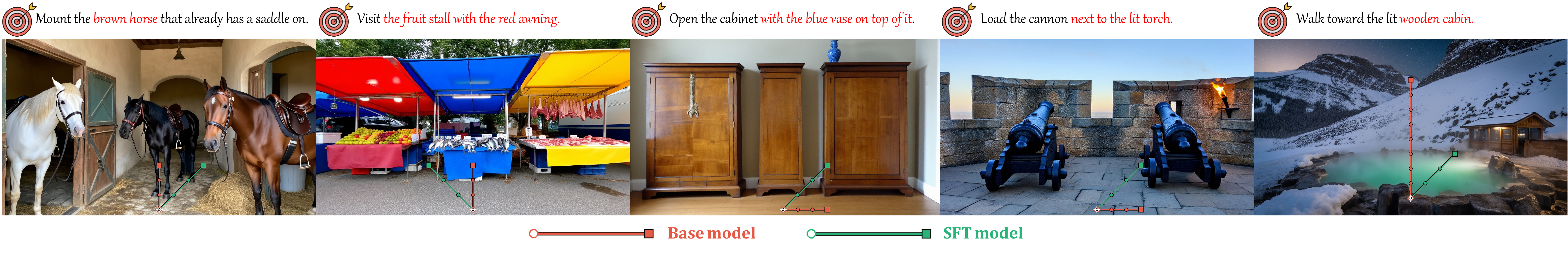}
\vspace{-1.5em}
    \caption{\textbf{Qualitative comparison examples on out-of-distribution
  scenes before and after Stage-I pretraining}, where the initial
  observations are synthesized by an image-generation model.
    }
    \vspace{-1em}
    \label{fig:case_visual_ood}
\end{figure}

\section{Experimental Results and Discussion}
\label{sec:eval}

Based on the two training stages introduced above, we report key implementation details and present the main empirical findings.

\subsection{EmbodiedWorld-200K pretraining substantially boosts embodied navigation-planning}


\begin{table}[t]
  \centering
  \small
  \caption{\textbf{Comparison on the EmbodiedWorld-200K evaluation split}. External baselines cover three families: VLN models adapted to our W/A/S/D action space, a game foundation model, and zero-shot frontier VLMs.}
  \label{tab:main_results_sft}
  \begin{tabular}{lcccc}
  \toprule
  \textbf{Model} & \textbf{TM $\uparrow$} & \textbf{DirAcc $\uparrow$} & \textbf{nDTW $\uparrow$} & \textbf{NE $\downarrow$} \\
  \midrule
  Uni-NaVid~\citep{zhang2024uninavid}   & 47.9 & 61.4 & 30.9 & 100.6 \\
  StreamVLN~\citep{wei2025streamvln}    & 47.2 & 61.5 & 25.9 & 114.7 \\
  NitroGen~\citep{magne2025nitrogen}    & 45.6 & 39.8 & 49.0 & \phantom{0}97.2 \\
  InternVL3-8B~\citep{zhu2025internvl3} & 45.3 & 57.4 & 63.4 & \phantom{0}59.7 \\
  Gemini-3.5~\citep{comanici2025gemini}   & 52.8 & 72.4 & 60.0 & \phantom{0}54.7 \\
  \midrule
  Qwen3-VL-4B (Base)~\citep{bai2025qwen3} & 54.9 & 70.1 & 67.1 & \phantom{0}51.7 \\
  \textbf{Qwen3-VL-4B (+SFT, ours)}     & 62.0 & 72.0 & 83.6 & \phantom{0}35.6 \\
  \bottomrule
  \end{tabular}
  \vspace{-1.6em}
  \end{table}

We finetune Qwen3-VL-4B~\cite{bai2025qwen3} with LoRA on the training
split of EmbodiedWorld-200K for 2 epochs. To position the resulting
model rather than only measure our own improvement, we further evaluate
three families of external baselines on the \emph{same} evaluation
split, with the \emph{same} metric toolkit and the \emph{same} prompt.
\textbf{(a) VLN models.} Uni-NaVid~\citep{zhang2024uninavid} and
StreamVLN~\citep{wei2025streamvln} are recent vision-and-language
navigation policies; we map their native action vocabularies onto our
W/A/S/D format one step per token.
\textbf{(b) A game foundation model.} NitroGen~\citep{magne2025nitrogen}
predicts gamepad actions from pixels after large-scale gameplay
pretraining; since it is not language-conditioned, it serves as an
instruction-free lower bound that isolates how much of the task is
solvable from visual reaction alone.
\textbf{(c) Zero-shot frontier VLMs.} The open-source
InternVL3-8B~\citep{zhu2025internvl3} and the proprietary Gemini-3.5 are
prompted with our baseline system prompt without any finetuning.
The last two rows of Tab.~\ref{tab:main_results_sft} are our own
backbone, Qwen3-VL-4B, before and after Stage-I
pretraining.

\begin{wrapfigure}{r}{0.45\textwidth}
\centering
\vspace{-1.6em}
\includegraphics[width=\linewidth]{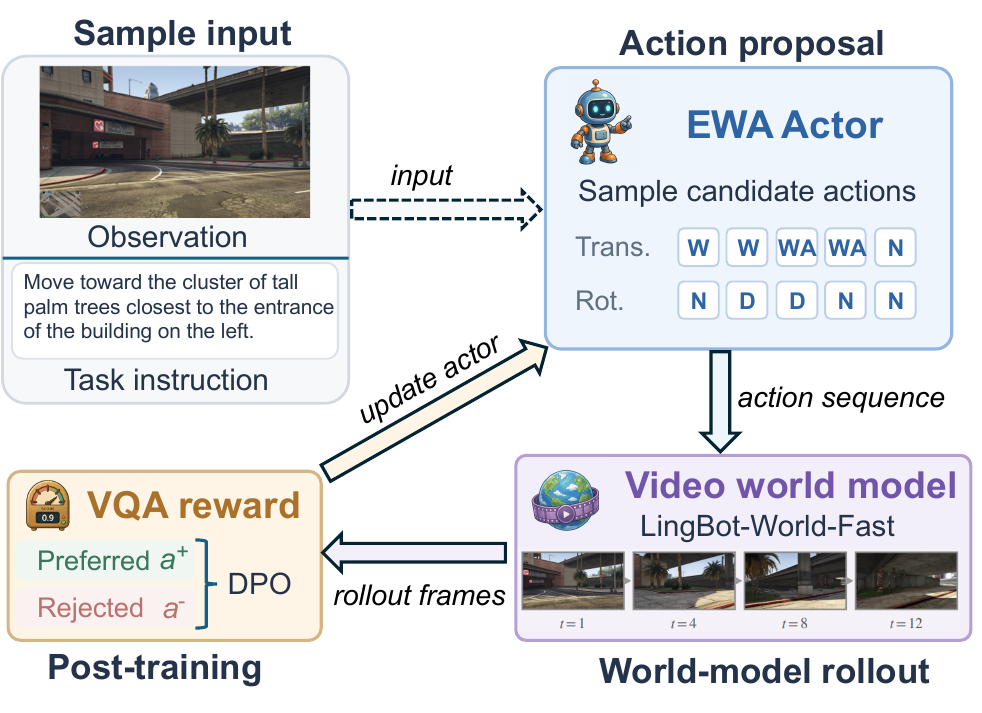}
\vspace{-1.6em}
\caption{\textbf{Overview of the Stage-II pipeline.} The actor proposes a candidate action sequence, the video world model rolls it out, and a VQA reward turns the rollout into a preferred/rejected pair for DPO.}
\vspace{-1.0em}
\label{fig:stage2_pipeline}
\end{wrapfigure}
Three observations follow. First, Stage-I SFT improves every metric over
the zero-shot backbone (TM $+7.1$, DirAcc $+1.9$, nDTW $+16.5$, NE
$-16.1$), confirming that the corpus injects the intended open-world
navigation-planning priors. Second, these gains are not obtainable from
a stronger general-purpose model: our model leads every external
baseline on TM, nDTW and NE ($+9.2$, $+20.2$ and $-19.1$ over the
strongest competitor) and matches the best DirAcc ($72.0$ vs.\ $72.4$);
notably, Gemini-3.5 attains that best DirAcc yet trails markedly on the
trajectory-level metrics, indicating that strong multimodal
understanding suffices to infer the coarse heading but not to reproduce
the goal-directed action sequence itself. Third, the benchmark is
discriminative rather than saturated: baseline scores span a wide range
(TM $45.3$--$54.9$, nDTW $25.9$--$67.1$, NE $51.7$--$114.7$), and the
instruction-free NitroGen records the lowest DirAcc ($39.8$, close to
chance), confirming that the task genuinely requires grounding the
language instruction rather than reacting to pixels alone.

In addition, Fig.~\ref{fig:vis_case_sft} shows qualitative examples on evaluation split, and Fig.~\ref{fig:case_visual_ood} further shows out-of-distribution samples whose initial observations $o_0$ are synthesized by an image-generation model~\citep{team2025zimage}. 
 We further analyze the effect of the Stage-I pretraining data scale in App.~\ref{app:stage1_scale}.
In summary, results show that pretraining on the proposed dataset successfully injects expected open-world embodied-planning priors for model.

\subsection{World-model-based RL post-training improves target-scene adaptation}

Stage I essentially fits the model to a broad collection of world environments and extracts the common knowledge shared across them; its capacity is therefore inherently bounded by the training distribution. To further adapt the model to a particular target environment — for example, to calibrate the precise physical magnitude that each unit action corresponds to under that environment's dynamics — closed-loop interaction with, and post-training on, that very environment is indispensable. This mirrors the common recipe in embodied-robotics research~\citep{kim2024openvla,black2024pi0,rt2}, where a model is first pretrained on broad and diverse corpora and then finetuned on a specific robot platform to adapt to its physical characteristics. Since directly building a real interactive platform is prohibitively expensive, we instead resort to a frontier video world model~\citep{team2026advancing} as a low-cost proxy. Concretely, we use LingBot-World-Fast~\citep{team2026advancing} as the interactive proxy and take
GTA5 as an illustrative target environment. GTA5 is an open-world game
widely used in academic research \citep{doan2018g2d,yun2021simulation}, yet it is not included in our training
distribution. We therefore treat it as a representative setting in which
large-scale sampling of real trajectories is costly.

\begin{wraptable}{r}{0.5\textwidth}
\centering
\vspace{-1.5em}
\fontsize{8.5pt}{12pt}\selectfont
\caption{Performance on GTA5 evaluation split, before (w/o DPO) and after (w/ DPO) Stage-II
  post-training.}
\vspace{-0.7em}
\label{tab:main_results_dpo}
\resizebox{0.5\textwidth}{!}{
\begin{tabular}{lcccc}
\toprule
\textbf{Setting} & \textbf{TM $\uparrow$} & \textbf{DirAcc $\uparrow$} & \textbf{nDTW $\uparrow$} & \textbf{NE $\downarrow$} \\
\midrule
w/o DPO  & 65.4 & 84.4 & 66.1 & 48.9 \\
w/ DPO  & 75.0 & 82.8 & 92.7 & 13.4 \\
\bottomrule
\end{tabular}

}
\vspace{-1em}
\end{wraptable}

Fig.~\ref{fig:stage2_pipeline} gives an overview of the resulting Stage-II pipeline: an initial observation and an automatically generated task instruction are fed to the actor, which samples a candidate action sequence; the sequence is executed inside the video world model to obtain the corresponding rollout frames, on which a VQA reward judges how well the instruction is fulfilled and yields a preferred/rejected action pair for DPO. 
We outline the key design choices below and defer the full
implementation to App.~\ref{app:stage2}.
Within this proxy, we align the physical magnitude that each unit control
action produces in the GTA5 environment with that produced inside
LingBot-World-Fast~\citep{team2026advancing}, and then initialize the
actor with the Stage-I SFT model rather than the base VLLM. We adopt the
SFT model as the actor because it has a basic embodied-planning ability,
so that its action sequences contain a non-trivial fraction of correct
decisions; combined with exploration, this yields rollouts of sufficient quality to form informative preference pairs for DPO~\citep{hu2026embeddingperturbed}. Starting from the
base model would produce rollouts too noisy to support effective
preference learning. We therefore roll out the SFT actor inside
LingBot-World-Fast~\citep{team2026advancing}, construct preference pairs
from the collected trajectories, and perform DPO post-training on the
actor. Tab.~\ref{tab:main_results_dpo} reports results on our
GTA5-specific evaluation split curated from the game. DPO post-training
on top of the Stage-I baseline significantly improves target-scene
adaptation.

\section{Conclusion}
\label{sec:conclusion}

In this work, we identified \textbf{Embodied World Agents (EWAs)}—general-purpose embodied agents that
can perceive, reason, and act across diverse open-world 3D environments—as an important yet
under-explored research direction, and contributed infrastructure support from both the \textbf{data}
and \textbf{training} perspectives. On the data side, we released \bench, a large-scale open-world
embodied-planning corpus of over 200K samples with multi-level annotations ranging from low-level
camera-motion trajectories to high-level semantic instructions, together with a unified evaluation split
and four task-specific metrics for fair and reproducible comparison. On the training side, we proposed
a two-stage training framework and instantiated a concrete baseline: offline supervised pretraining on
\bench followed by interactive RL post-training that leverages a frontier video world model as a
low-cost environment proxy.
Empirically, offline pretraining on \bench substantially strengthens open-world embodied planning       
ability across all four evaluation metrics, while world-model-based interactive post-training further
improves adaptation to out-of-distribution scenes. 
We release the dataset, checkpoints, and code as a shared foundation for EWA research.

\section*{Acknowledgments and Disclosure of Funding}
\label{sec:ack}
This work was conducted during an internship at Tencent Hunyuan, and was supported by Tencent Hunyuan.


\bibliographystyle{unsrtnat}
\bibliography{references}

@article{hurst2024gpt,
  title={Gpt-4o system card},
  author={Hurst, Aaron and Lerer, Adam and Goucher, Adam P and Perelman, Adam and Ramesh, Aditya and Clark, Aidan and Ostrow, AJ and Welihinda, Akila and Hayes, Alan and Radford, Alec and others},
  journal={arXiv preprint arXiv:2410.21276},
  year={2024}
}

@article{anthropic2024claude,
  title={The claude 3 model family: Opus, sonnet, haiku},
  author={Anthropic, AI},
  journal={Claude-3 Model Card},
  volume={1},
  number={1},
  pages={4},
  year={2024}
}

@article{comanici2025gemini,
  title={{Gemini 2.5}: Pushing the Frontier with Advanced Reasoning, Multimodality, Long Context, and Next Generation Agentic Capabilities},
  author={Comanici, Gheorghe and Bieber, Eric and Schaekermann, Mike and Pasupat, Ice and Sachdeva, Noveen and Dhillon, Inderjit and Blistein, Marcel and Ram, Ori and Zhang, Dan and Rosen, Evan and others},
  journal={arXiv preprint arXiv:2507.06261},
  year={2025}
}

@article{bai2025qwen3,
  title={{Qwen3-VL} Technical Report},
  author={Bai, Shuai and Cai, Yuxuan and Chen, Ruizhe and Chen, Keqin and Chen, Xionghui and Cheng, Zesen and Deng, Lianghao and Ding, Wei and Gao, Chang and Ge, Chunjiang and others},
  journal={arXiv preprint arXiv:2511.21631},
  year={2025}
}

@article{wang2025internvl3,
  title={{InternVL3.5}: Advancing Open-Source Multimodal Models in Versatility, Reasoning, and Efficiency},
  author={Wang, Weiyun and Gao, Zhangwei and Gu, Lixin and Pu, Hengjun and Cui, Long and Wei, Xingguang and Liu, Zhaoyang and Jing, Linglin and Ye, Shenglong and Shao, Jie and others},
  journal={arXiv preprint arXiv:2508.18265},
  year={2025}
}

@inproceedings{majumdar2024openeqa,
  title={{OpenEQA}: Embodied Question Answering in the Era of Foundation Models},
  author={Majumdar, Arjun and Ajay, Anurag and Zhang, Xiaohan and Putta, Pranav and Yenamandra, Sriram and Henaff, Mikael and Silwal, Sneha and Mcvay, Paul and Maksymets, Oleksandr and Arnaud, Sergio and others},
  booktitle={Proceedings of the IEEE/CVF Conference on Computer Vision and Pattern Recognition (CVPR)},
  pages={16488--16498},
  year={2024}
}

@article{fung2025embodied,
  title={Embodied {AI} Agents: Modeling the World},
  author={Fung, Pascale and Bachrach, Yoram and Celikyilmaz, Asli and Chaudhuri, Kamalika and Chen, Delong and Chung, Willy and Dupoux, Emmanuel and Gong, Hongyu and J{\'e}gou, Herv{\'e} and Lazaric, Alessandro and others},
  journal={arXiv preprint arXiv:2506.22355},
  year={2025}
}

@article{huang2025vipe,
  title={Vipe: Video pose engine for 3d geometric perception},
  author={Huang, Jiahui and Zhou, Qunjie and Rabeti, Hesam and Korovko, Aleksandr and Ling, Huan and Ren, Xuanchi and Shen, Tianchang and Gao, Jun and Slepichev, Dmitry and Lin, Chen-Hsuan and others},
  journal={arXiv preprint arXiv:2508.10934},
  year={2025}
}

@article{rafailov2023dpo,
  title={Direct preference optimization: Your language model is secretly a reward model},
  author={Rafailov, Rafael and Sharma, Archit and Mitchell, Eric and Manning, Christopher D and Ermon, Stefano and Finn, Chelsea},
  journal={Advances in neural information processing systems},
  volume={36},
  pages={53728--53741},
  year={2023}
}

@article{anderson2018evaluation,
  title={On evaluation of embodied navigation agents},
  author={Anderson, Peter and Chang, Angel and Chaplot, Devendra Singh and Dosovitskiy, Alexey and Gupta, Saurabh and Koltun, Vladlen and Kosecka, Jana and Malik, Jitendra and Mottaghi, Roozbeh and Savva, Manolis and others},
  journal={arXiv preprint arXiv:1807.06757},
  year={2018}
}

@inproceedings{krantz2020beyond,
  title={Beyond the nav-graph: Vision-and-language navigation in continuous environments},
  author={Krantz, Jacob and Wijmans, Erik and Majumdar, Arjun and Batra, Dhruv and Lee, Stefan},
  booktitle={European Conference on Computer Vision},
  pages={104--120},
  year={2020},
  organization={Springer}
}

@inproceedings{anderson2018vision,
  title={Vision-and-language navigation: Interpreting visually-grounded navigation instructions in real environments},
  author={Anderson, Peter and Wu, Qi and Teney, Damien and Bruce, Jake and Johnson, Mark and S{\"u}nderhauf, Niko and Reid, Ian and Gould, Stephen and Van Den Hengel, Anton},
  booktitle={Proceedings of the IEEE conference on computer vision and pattern recognition},
  pages={3674--3683},
  year={2018}
}

@misc{genie3,
  title        = {Genie 3: A New Frontier for World Models},
  author       = {{Google DeepMind}},
  year         = {2025},
  howpublished = {\url{https://deepmind.google/blog/genie-3-a-new-frontier-for-world-models/}}
}

@article{cosmos2025,
  title={Cosmos world foundation model platform for physical ai},
  author={Agarwal, Niket and Ali, Arslan and Bala, Maciej and Balaji, Yogesh and Barker, Erik and Cai, Tiffany and Chattopadhyay, Prithvijit and Chen, Yongxin and Cui, Yin and Ding, Yifan and others},
  journal={arXiv preprint arXiv:2501.03575},
  year={2025}
}

@article{nvidia2025cosmostransfer1conditionalworldgeneration,
  title={Cosmos-transfer1: Conditional world generation with adaptive multimodal control},
  author={Alhaija, Hassan Abu and Alvarez, Jose and Bala, Maciej and Cai, Tiffany and Cao, Tianshi and Cha, Liz and Chen, Joshua and Chen, Mike and Ferroni, Francesco and Fidler, Sanja and others},
  journal={arXiv preprint arXiv:2503.14492},
  year={2025}
}

@article{wang2024jarvis,
  title={Jarvis-1: Open-world multi-task agents with memory-augmented multimodal language models},
  author={Wang, Zihao and Cai, Shaofei and Liu, Anji and Jin, Yonggang and Hou, Jinbing and Zhang, Bowei and Lin, Haowei and He, Zhaofeng and Zheng, Zilong and Yang, Yaodong and others},
  journal={IEEE Transactions on Pattern Analysis and Machine Intelligence},
  volume={47},
  number={3},
  pages={1894--1907},
  year={2024},
  publisher={IEEE}
}

@article{tan2024cradle,
  title={Cradle: Empowering foundation agents towards general computer control},
  author={Tan, Weihao and Zhang, Wentao and Xu, Xinrun and Xia, Haochong and Ding, Ziluo and Li, Boyu and Zhou, Bohan and Yue, Junpeng and Jiang, Jiechuan and Li, Yewen and others},
  journal={arXiv preprint arXiv:2403.03186},
  year={2024}
}

@article{yang2023learning,
  title={Learning interactive real-world simulators},
  author={Yang, Sherry and Du, Yilun and Ghasemipour, Kamyar and Tompson, Jonathan and Kaelbling, Leslie and Schuurmans, Dale and Abbeel, Pieter},
  journal={arXiv preprint arXiv:2310.06114},
  year={2023}
}

@article{hu2022lora,
  title={Lora: Low-rank adaptation of large language models.},
  author={Hu, Edward J and Shen, Yelong and Wallis, Phillip and Allen-Zhu, Zeyuan and Li, Yuanzhi and Wang, Shean and Wang, Liang and Chen, Weizhu and others},
  journal={Iclr},
  volume={1},
  number={2},
  pages={3},
  year={2022}
}

@article{cheng2025embodiedeval,
  title={Embodiedeval: Evaluate multimodal llms as embodied agents},
  author={Cheng, Zhili and Tu, Yuge and Li, Ran and Dai, Shiqi and Hu, Jinyi and Hu, Shengding and Li, Jiahao and Shi, Yang and Yu, Tianyu and Chen, Weize and others},
  journal={arXiv preprint arXiv:2501.11858},
  year={2025}
}

@inproceedings{zhu2025understand3dscenebridging,
  title={Move to understand a 3d scene: Bridging visual grounding and exploration for efficient and versatile embodied navigation},
  author={Zhu, Ziyu and Wang, Xilin and Li, Yixuan and Zhang, Zhuofan and Ma, Xiaojian and Chen, Yixin and Jia, Baoxiong and Liang, Wei and Yu, Qian and Deng, Zhidong and others},
  booktitle={Proceedings of the IEEE/CVF International Conference on Computer Vision},
  pages={8120--8132},
  year={2025}
}

@article{matrixgame2025,
  title={Matrix-game 2.0: An open-source real-time and streaming interactive world model},
  author={He, Xianglong and Peng, Chunli and Liu, Zexiang and Wang, Boyang and Zhang, Yifan and Cui, Qi and Kang, Fei and Jiang, Biao and An, Mengyin and Ren, Yangyang and others},
  journal={arXiv preprint arXiv:2508.13009},
  year={2025}
}

@article{hyworld2025,
  title={Hunyuanworld 1.0: Generating immersive, explorable, and interactive 3d worlds from words or pixels},
  author={Team, HunyuanWorld and Wang, Zhenwei and Liu, Yuhao and Wu, Junta and Gu, Zixiao and Wang, Haoyuan and Zuo, Xuhui and Huang, Tianyu and Li, Wenhuan and Zhang, Sheng and others},
  journal={arXiv preprint arXiv:2507.21809},
  year={2025}
}

@article{pan2025,
  title={Pan: A world model for general, interactable, and long-horizon world simulation},
  author={Xiang, Jiannan and Gu, Yi and Liu, Zihan and Feng, Zeyu and Gao, Qiyue and Hu, Yiyan and Huang, Benhao and Liu, Guangyi and Yang, Yichi and Zhou, Kun and others},
  journal={arXiv preprint arXiv:2511.09057},
  year={2025}
}

@article{che2024gamegen,
  title={Gamegen-x: Interactive open-world game video generation},
  author={Che, Haoxuan and He, Xuanhua and Liu, Quande and Jin, Cheng and Chen, Hao},
  journal={arXiv preprint arXiv:2411.00769},
  year={2024}
}

@inproceedings{du2023videolanguageplanning,
  title={Video language planning},
  author={Du, Yilun and Yang, Sherry and Florence, Pete and Xia, Fei and Wahid, Ayzaan and Sermanet, Pierre and Yu, Tianhe and Abbeel, Pieter and Tenenbaum, Joshua B and Kaelbling, Leslie Pack and others},
  booktitle={The Twelfth International Conference on Learning Representations},
  year={2024}
}

@article{zhou2024robodreamer,
  title={Robodreamer: Learning compositional world models for robot imagination},
  author={Zhou, Siyuan and Du, Yilun and Chen, Jiaben and Li, Yandong and Yeung, Dit-Yan and Gan, Chuang},
  journal={arXiv preprint arXiv:2404.12377},
  year={2024}
}

@article{du2023unipi,
  title={Learning universal policies via text-guided video generation},
  author={Du, Yilun and Yang, Sherry and Dai, Bo and Dai, Hanjun and Nachum, Ofir and Tenenbaum, Josh and Schuurmans, Dale and Abbeel, Pieter},
  journal={Advances in neural information processing systems},
  volume={36},
  pages={9156--9172},
  year={2023}
}

@article{kim2024openvla,
  title={Openvla: An open-source vision-language-action model},
  author={Kim, Moo Jin and Pertsch, Karl and Karamcheti, Siddharth and Xiao, Ted and Balakrishna, Ashwin and Nair, Suraj and Rafailov, Rafael and Foster, Ethan and Lam, Grace and Sanketi, Pannag and others},
  journal={arXiv preprint arXiv:2406.09246},
  year={2024}
}

@article{black2024pi0,
  title={$\pi_0$: A Vision-Language-Action Flow Model for General Robot Control},
  author={Black, Kevin and Brown, Noah and Driess, Danny and Esmail, Adnan and Equi, Michael and Finn, Chelsea and Fusai, Niccolo and Groom, Lachy and Hausman, Karol and Ichter, Brian and others},
  journal={arXiv preprint arXiv:2410.24164},
  year={2024}
}

@article{intelligence2504pi0,
  title={$\pi_{0.5}$: a Vision-Language-Action Model with Open-World Generalization},
  author={Intelligence, Physical and Black, Kevin and Brown, Noah and Darpinian, James and Dhabalia, Karan and Driess, Danny and Esmail, Adnan and Equi, Michael and Finn, Chelsea and Fusai, Niccolo and others},
  journal={arXiv preprint arXiv:2504.16054},
  year={2025}
}

@inproceedings{baker2022video,
  title={Video pretraining ({VPT}): Learning to act by watching unlabeled online videos},
  author={Baker, Bowen and Akkaya, Ilge and Zhokov, Peter and Huizinga, Joost and Tang, Jie and Ecoffet, Adrien and Houghton, Brandon and Sampedro, Raul and Clune, Jeff},
  booktitle={Advances in Neural Information Processing Systems},
  year={2022}
}

@article{bytedance2025lumine,
  title={{Lumine: An Open Recipe for Building Generalist Agents
in 3D Open Worlds}},
  author={{ByteDance Seed} and Tan, Weihao and Li, Xiangyang and Fang, Yunhao and Yao, Heyuan and Yan, Shi and Luo, Hao and Ao, Tenglong and Li, Huihui and Ren, Hongbin and Yi, Bairen and Qin, Yujia and An, Bo and Liu, Libin and Shi, Guang},
  journal={arXiv preprint arXiv:2511.08892},
  year={2025}
}

@inproceedings{lee2022multi,
  title={Multi-game decision transformers},
  author={Lee, Kuang-Huei and Nachum, Ofir and Yang, Mengjiao Sherry and Lee, Lisa and Freeman, Daniel and Guadarrama, Sergio and Fischer, Ian and Xu, Winnie and Jang, Eric and Michalewski, Henryk and others},
  booktitle={Advances in Neural Information Processing Systems},
  year={2022}
}

@misc{gamecraft1,
    title={Hunyuan-GameCraft: High-dynamic Interactive Game Video Generation with Hybrid History Condition}, 
    author={Jiaqi Li and Junshu Tang and Zhiyong Xu and Longhuang Wu and Yuan Zhou and Shuai Shao and Tianbao Yu and Zhiguo Cao and Qinglin Lu},
    year={2025},
    eprint={2506.17201},
    archivePrefix={arXiv},
    primaryClass={cs.CV},
    url={https://arxiv.org/abs/2506.17201}, 
}

@misc{gamecraft2,
    title={Hunyuan-GameCraft-2: Instruction-following Interactive Game World Model}, 
    author={Junshu Tang and Jiacheng Liu and Jiaqi Li and Longhuang Wu and Haoyu Yang and Penghao Zhao and Siruis Gong and Xiang Yuan and Shuai Shao and Qinglin Lu},
    year={2025},
    eprint={2511.23429},
    archivePrefix={arXiv},
    primaryClass={cs.CV},
    url={https://arxiv.org/abs/2511.23429}, 
}

@article{sima1,
  title={Scaling instructable agents across many simulated worlds},
  author={{SIMA Team} and Raad, Maria Abi and Ahuja, Arun and Barros, Catarina and Besse, Frederic and Bolt, Andrew and Bolton, Adrian and Brownfield, Bethanie and Buttimore, Gavin and Cant, Max and Chakera, Sarah and others},
  journal={arXiv preprint arXiv:2404.10179},
  year={2024}
}

@article{sima2,
  title={Sima 2: A generalist embodied agent for virtual worlds},
  author={Bolton, Adrian and Lerchner, Alexander and Cordell, Alexandra and Moufarek, Alexandre and Bolt, Andrew and Lampinen, Andrew and Mitenkova, Anna and Hallingstad, Arne Olav and Vujatovic, Bojan and Li, Bonnie and others},
  journal={arXiv preprint arXiv:2512.04797},
  year={2025}
}

@inproceedings{kalashnikov2018scalable,
  title={Scalable deep reinforcement learning for vision-based robotic manipulation},
  author={Kalashnikov, Dmitry and Irpan, Alex and Pastor, Peter and Ibarz, Julian and Herzog, Alexander and Jang, Eric and Quillen, Deirdre and Holly, Ethan and Kalakrishnan, Mrinal and Vanhoucke, Vincent and others},
  booktitle={Conference on robot learning},
  pages={651--673},
  year={2018},
  organization={PMLR}
}

@inproceedings{o2024open,
  title={Open x-embodiment: Robotic learning datasets and rt-x models: Open x-embodiment collaboration 0},
  author={O’Neill, Abby and Rehman, Abdul and Maddukuri, Abhiram and Gupta, Abhishek and Padalkar, Abhishek and Lee, Abraham and Pooley, Acorn and Gupta, Agrim and Mandlekar, Ajay and Jain, Ajinkya and others},
  booktitle={2024 IEEE International Conference on Robotics and Automation (ICRA)},
  pages={6892--6903},
  year={2024},
  organization={IEEE}
}

@inproceedings{walke2023bridgedata,
    title={BridgeData V2: A Dataset for Robot Learning at Scale},
    author={Walke, Homer and Black, Kevin and Lee, Abraham and Kim, Moo Jin and Du, Max and Zheng, Chongyi and Zhao, Tony and Hansen-Estruch, Philippe and Vuong, Quan and He, Andre and Myers, Vivek and Fang, Kuan and Finn, Chelsea and Levine, Sergey},
    booktitle={Conference on Robot Learning (CoRL)},
    year={2023}
}

@article{team2024octo,
  title={Octo: An open-source generalist robot policy},
  author={Team, Octo Model and Ghosh, Dibya and Walke, Homer and Pertsch, Karl and Black, Kevin and Mees, Oier and Dasari, Sudeep and Hejna, Joey and Kreiman, Tobias and Xu, Charles and others},
  journal={arXiv preprint arXiv:2405.12213},
  year={2024}
}

@inproceedings{openxembodiment2023,
  title={Open x-embodiment: Robotic learning datasets and rt-x models: Open x-embodiment collaboration 0},
  author={O’Neill, Abby and Rehman, Abdul and Maddukuri, Abhiram and Gupta, Abhishek and Padalkar, Abhishek and Lee, Abraham and Pooley, Acorn and Gupta, Agrim and Mandlekar, Ajay and Jain, Ajinkya and others},
  booktitle={2024 IEEE International Conference on Robotics and Automation (ICRA)},
  pages={6892--6903},
  year={2024},
  organization={IEEE}
}

@inproceedings{rt2,
  title={Rt-2: Vision-language-action models transfer web knowledge to robotic control},
  author={Zitkovich, Brianna and Yu, Tianhe and Xu, Sichun and Xu, Peng and Xiao, Ted and Xia, Fei and Wu, Jialin and Wohlhart, Paul and Welker, Stefan and Wahid, Ayzaan and others},
  booktitle={Conference on Robot Learning},
  pages={2165--2183},
  year={2023},
  organization={PMLR}
}

@article{driess2023palme,
  title={Palm-e: An embodied multimodal language model},
  author={Driess, Danny and Xia, Fei and Sajjadi, Mehdi SM and Lynch, Corey and Chowdhery, Aakanksha and Ichter, Brian and Wahid, Ayzaan and Tompson, Jonathan and Vuong, Quan and Yu, Tianhe and others},
  journal={arXiv preprint arXiv:2303.03378},
  year={2023}
}

@article{saycan2022,
  title={Do as i can, not as i say: Grounding language in robotic affordances},
  author={Ahn, Michael and Brohan, Anthony and Brown, Noah and Chebotar, Yevgen and Cortes, Omar and David, Byron and Finn, Chelsea and Fu, Chuyuan and Gopalakrishnan, Keerthana and Hausman, Karol and others},
  journal={arXiv preprint arXiv:2204.01691},
  year={2022}
}

@article{fan2022minedojo,
  title={Minedojo: Building open-ended embodied agents with internet-scale knowledge},
  author={Fan, Linxi and Wang, Guanzhi and Jiang, Yunfan and Mandlekar, Ajay and Yang, Yuncong and Zhu, Haoyi and Tang, Andrew and Huang, De-An and Zhu, Yuke and Anandkumar, Anima},
  journal={Advances in Neural Information Processing Systems},
  volume={35},
  pages={18343--18362},
  year={2022}
}

@article{wang2023voyager,
  title={Voyager: An open-ended embodied agent with large language models},
  author={Wang, Guanzhi and Xie, Yuqi and Jiang, Yunfan and Mandlekar, Ajay and Xiao, Chaowei and Zhu, Yuke and Fan, Linxi and Anandkumar, Anima},
  journal={arXiv preprint arXiv:2305.16291},
  year={2023}
}

@article{lifshitz2023steve,
  title={Steve-1: A generative model for text-to-behavior in minecraft},
  author={Lifshitz, Shalev and Paster, Keiran and Chan, Harris and Ba, Jimmy and McIlraith, Sheila},
  journal={Advances in Neural Information Processing Systems},
  volume={36},
  pages={69900--69929},
  year={2023}
}

@article{hafner2023dreamerv3,
  title={Mastering diverse domains through world models},
  author={Hafner, Danijar and Pasukonis, Jurgis and Ba, Jimmy and Lillicrap, Timothy},
  journal={arXiv preprint arXiv:2301.04104},
  year={2023}
}

@article{micheli2023iris,
  title={Transformers are sample-efficient world models},
  author={Micheli, Vincent and Alonso, Eloi and Fleuret, Fran{\c{c}}ois},
  journal={arXiv preprint arXiv:2209.00588},
  year={2022}
}

@inproceedings{bruce2024genie,
  title={Genie: Generative interactive environments},
  author={Bruce, Jake and Dennis, Michael D and Edwards, Ashley and Parker-Holder, Jack and Shi, Yuge and Hughes, Edward and Lai, Matthew and Mavalankar, Aditi and Steigerwald, Richie and Apps, Chris and others},
  booktitle={Forty-first International Conference on Machine Learning},
  year={2024}
}

@article{valevski2024gamengen,
  title={Diffusion models are real-time game engines},
  author={Valevski, Dani and Leviathan, Yaniv and Arar, Moab and Fruchter, Shlomi},
  journal={arXiv preprint arXiv:2408.14837},
  year={2024}
}

@article{alonso2024diamond,
  title={Diffusion for world modeling: Visual details matter in atari},
  author={Alonso, Eloi and Jelley, Adam and Micheli, Vincent and Kanervisto, Anssi and Storkey, Amos and Pearce, Tim and Fleuret, Fran{\c{c}}ois},
  journal={Advances in Neural Information Processing Systems},
  volume={37},
  pages={58757--58791},
  year={2024}
}

@article{hu2023gaia1,
  title={Gaia-1: A generative world model for autonomous driving},
  author={Hu, Anthony and Russell, Lloyd and Yeo, Hudson and Murez, Zak and Fedoseev, George and Kendall, Alex and Shotton, Jamie and Corrado, Gianluca},
  journal={arXiv preprint arXiv:2309.17080},
  year={2023}
}

@article{mazzaglia2024genrl,
  title={Genrl: Multimodal-foundation world models for generalization in embodied agents},
  author={Mazzaglia, Pietro and Verbelen, Tim and Dhoedt, Bart and Courville, Aaron and Rajeswar, Sai},
  journal={Advances in neural information processing systems},
  volume={37},
  pages={27529--27555},
  year={2024}
}

@article{cheang2024gr2,
  title={Gr-2: A generative video-language-action model with web-scale knowledge for robot manipulation},
  author={Cheang, Chi-Lam and Chen, Guangzeng and Jing, Ya and Kong, Tao and Li, Hang and Li, Yifeng and Liu, Yuxiao and Wu, Hongtao and Xu, Jiafeng and Yang, Yichu and others},
  journal={arXiv preprint arXiv:2410.06158},
  year={2024}
}

@article{silver2025era,
  title={Welcome to the era of experience},
  author={Silver, David and Sutton, Richard S},
  journal={Google AI},
  volume={1},
  pages={11},
  year={2025}
}

@article{rocamonde2024vlmrm,
  title={Vision-language models are zero-shot reward models for reinforcement learning},
  author={Rocamonde, Juan and Montesinos, Victoriano and Nava, Elvis and Perez, Ethan and Lindner, David},
  journal={arXiv preprint arXiv:2310.12921},
  year={2023}
}

@article{wang2024rlvlmf,
  title={Rl-vlm-f: Reinforcement learning from vision language foundation model feedback},
  author={Wang, Yufei and Sun, Zhanyi and Zhang, Jesse and Xian, Zhou and Biyik, Erdem and Held, David and Erickson, Zackory},
  journal={arXiv preprint arXiv:2402.03681},
  year={2024}
}

@article{li2024cogact,
  title={Cogact: A foundational vision-language-action model for synergizing cognition and action in robotic manipulation},
  author={Li, Qixiu and Liang, Yaobo and Wang, Zeyu and Luo, Lin and Chen, Xi and Liao, Mozheng and Wei, Fangyun and Deng, Yu and Xu, Sicheng and Zhang, Yizhong and others},
  journal={arXiv preprint arXiv:2411.19650},
  year={2024}
}

@article{hou2024diffusion,
  title={Diffusion transformer policy},
  author={Hou, Zhi and Zhang, Tianyi and Xiong, Yuwen and Pu, Hengjun and Zhao, Chengyang and Tong, Ronglei and Qiao, Yu and Dai, Jifeng and Chen, Yuntao},
  journal={arXiv preprint arXiv:2410.15959},
  year={2024}
}

@article{sapkota2025vision,
  title={Vision-language-action models: Concepts, progress, applications and challenges},
  author={Sapkota, Ranjan and Cao, Yang and Roumeliotis, Konstantinos I and Karkee, Manoj},
  journal={arXiv preprint arXiv:2505.04769},
  year={2025}
}

@article{team2026advancing,
  title={Advancing Open-source World Models},
  author={Team, Robbyant and Gao, Zelin and Wang, Qiuyu and Zeng, Yanhong and Zhu, Jiapeng and Cheng, Ka Leong and Li, Yixuan and Wang, Hanlin and Xu, Yinghao and Ma, Shuailei and others},
  journal={arXiv preprint arXiv:2601.20540},
  year={2026}
}

@article{team2025zimage,
  title={Z-Image: An Efficient Image Generation Foundation Model with Single-Stream Diffusion Transformer},
  author={Z-Image Team},
  journal={arXiv preprint arXiv:2511.22699},
  year={2025}
}

@inproceedings{yun2021simulation,
  title={Simulation of self-driving system by implementing digital twin with GTA5},
  author={Yun, Heuijee and Park, Daejin},
  booktitle={2021 international conference on electronics, information, and communication (ICEIC)},
  pages={1--2},
  year={2021},
  organization={IEEE}
}

@article{doan2018g2d,
  title={G2d: from gta to data},
  author={Doan, Anh-Dzung and Jawaid, Abdul Mohsi and Do, Thanh-Toan and Chin, Tat-Jun},
  journal={arXiv preprint arXiv:1806.07381},
  year={2018}
}

@article{zhang2024uninavid,
  title={Uni-navid: A video-based vision-language-action model for unifying embodied navigation tasks},
  author={Zhang, Jiazhao and Wang, Kunyu and Wang, Shaoan and Li, Minghan and Liu, Haoran and Wei, Songlin and Wang, Zhongyuan and Zhang, Zhizheng and Wang, He},
  journal={arXiv preprint arXiv:2412.06224},
  year={2024}
}

@inproceedings{wei2025streamvln,
  title={Streamvln: Streaming vision-and-language navigation via slowfast context modeling},
  author={Wei, Meng and Wan, Chenyang and Yu, Xiqian and Wang, Tai and Mao, Xiaohan and Zhu, Chenming and Cai, Wenzhe and Wang, Hanqing and Chen, Yilun and Liu, Xihui and others},
  booktitle={2026 IEEE International Conference on Robotics and Automation (ICRA)},
  pages={22174--22181},
  year={2026},
  organization={IEEE}
}

@misc{magne2025nitrogen,
  title={NitroGen: A Foundation Model for Generalist Gaming Agents},
  author={Magne, L and Awadalla, A and Wang, G and Xu, Y and Belofsky, J and Hu, F and Kim, J and Schmidt, L and Gkioxari, G and Kautz, J and others}
}

@article{zhu2025internvl3,
  title={Internvl3: Exploring advanced training and test-time recipes for open-source multimodal models},
  author={Zhu, Jinguo and Wang, Weiyun and Chen, Zhe and Liu, Zhaoyang and Ye, Shenglong and Gu, Lixin and Tian, Hao and Duan, Yuchen and Su, Weijie and Shao, Jie and others},
  journal={arXiv preprint arXiv:2504.10479},
  year={2025}
}

@misc{hu2026blameerrantwords,
      title={Blame Errant Words, Not the Canvas: Defect-Aware Spatial Credit Assignment for Visual Text Generation}, 
      author={Sujie Hu and Huaisong Zhang and Pujing Yang and Qian Wang and Xiaomin Li and Xinhang Leng and Chen Li and Jing Lyu and Xiu Li and Ling Pan},
      year={2026},
      eprint={2610.20794},
      archivePrefix={arXiv},
      primaryClass={cs.CV},
      url={https://arxiv.org/abs/2610.20794}, 
}

@inproceedings{
    hu2026embeddingperturbed,
    title={Embedding-perturbed Exploration Preference Optimization for Flow Models},
    author={Sujie Hu and Chubin Chen and Jiashu Zhu and Jiahong Wu and Xiangxiang Chu and Xiu Li},
    booktitle={Forty-third International Conference on Machine Learning},
    year={2026},
    url={https://openreview.net/forum?id=Ai4vdy80p5}
}

@inproceedings{chen2026taming,
  title={Taming preference mode collapse via directional decoupling alignment in diffusion reinforcement learning},
  author={Chen, Chubin and Hu, Sujie and Zhu, Jiashu and Wu, Meiqi and Chen, Jintao and Li, Yanxun and Huang, Nisha and Fang, Chengyu and Wu, Jiahong and Chu, Xiangxiang and others},
  booktitle={Proceedings of the IEEE/CVF Conference on Computer Vision and Pattern Recognition},
  pages={12775--12786},
  year={2026}
}

@article{feng2026enhancing,
  title={Enhancing Train-Free Infinite-Frame Generation for Consistent Long Videos},
  author={Feng, Xiaokun and Zhu, Jiashu and Wu, Meiqi and Chen, Chubin and Mao, Fangyuan and Guo, Haiyang and Wu, Jiahong and Chu, Xiangxiang and Huang, Kaiqi},
  journal={arXiv preprint arXiv:2605.18233},
  year={2026}
}

@inproceedings{feng2026narrlv,
  title={Narrlv: Towards a comprehensive narrative-centric evaluation for long video generation},
  author={Feng, Xiaokun and Yu, Haiming and Wu, Meiqi and Hu, Shiyu and Chen, Jintao and Zhu, Chen and Wu, Jiahong and Chu, Xiangxiang and Huang, Kaiqi},
  booktitle={International Conference on Learning Representations},
  volume={2026},
  pages={20966--21000},
  year={2026}
}

@inproceedings{ling2025vmbench,
  title={Vmbench: A benchmark for perception-aligned video motion generation},
  author={Ling, Xinran and Zhu, Chen and Wu, Meiqi and Li, Hangyu and Feng, Xiaokun and Yang, Cundian and Hao, Aiming and Zhu, Jiashu and Wu, Jiahong and Chu, Xiangxiang},
  booktitle={2025 IEEE/CVF International Conference on Computer Vision (ICCV)},
  pages={13087--13098},
  year={2025},
  organization={IEEE}
}


\newpage
\appendix

\setcounter{figure}{0}
\setcounter{table}{0}
\renewcommand{\thefigure}{A\arabic{figure}}
\renewcommand{\thetable}{A\arabic{table}}

\section{More Details About the EmbodiedWorld-200K Construction Pipeline}
\label{app:dataset}

This appendix complements Sec.~\ref{sec:dataset}.2 and provides the full specification of the three annotation steps that turn a raw gameplay clip into a triplet $(o_0, \ell, \mathbf{a}_{1:T})$: (i) navigation-coherent segmentation, (ii) magnitude-based variable-length W/A/S/D discretization, and (iii) instruction annotation with verification.

\subsection{Navigation-Coherent Segmentation}
\label{app:ncs}

We start from the 6-DoF camera trajectory recovered by VIPE~\citep{huang2025vipe}. For each clip, we first re-express the per-frame poses in a local coordinate frame anchored at the first frame and smooth the resulting signal with a sliding-window average of size $W_{\text{smooth}} = 5$ to suppress high-frequency jitter.

\paragraph{Principal-direction tracking and cut criterion.}
On the smoothed signal we maintain a running estimate of the cumulative principal motion direction. Whenever the instantaneous motion deviates from this principal direction by more than $\theta_{\text{cut}} = 90^\circ$ for at least $F_{\text{cut}} = 3$ consecutive frames, a cut is triggered; the portion between two adjacent cuts is treated as one candidate navigation-coherent segment (NCS). Intuitively, this criterion ensures that every NCS corresponds to a temporally continuous interval during which the agent moves towards a stable destination.

\paragraph{Quality filters.}
The coarse segmentation above is followed by two filters to discard erratic or under-informative candidates:
\begin{itemize}
    \item \textbf{Displacement-vs-path-length ratio.} For each candidate segment we compute the ratio of its net displacement (start-to-end straight-line distance) to the integrated path length. Candidates with a ratio below $\rho_{\text{disp}} = 0.3$, or whose curvature variance exceeds $\sigma_\kappa^2 = 1.5$, are dropped as overly meandering or shaky.
    \item \textbf{Minimum-length constraint.} Any remaining candidate shorter than $L_{\text{min}} = 60$ frames is dropped, so that each surviving NCS contains enough visual context for the downstream annotation model to reason about the navigation goal.
\end{itemize}

\paragraph{Initial observation $o_0$ and frame-index alignment.}
The first frame of every surviving NCS is exported as the initial visual observation $o_0$ of the corresponding EWA sample. When the recovered-pose sampling rate disagrees with the video sampling rate, we remap the pose-index of the first frame to the corresponding video-index before exporting, so that $o_0$ is strictly aligned with the first pose used by the NCS.

\subsection{Magnitude-Based Variable-Length W/A/S/D Discretization}
\label{app:quant}

As stated in Sec.~\ref{sec:dataset}.2 Step~2, the central design question here is how to calibrate the physical amount of motion that each W/A/S/D token represents: too coarse a unit yields too-short sequences that cannot carry fine-grained navigation signals, whereas too fine a unit produces overly long sequences that are hard to learn. Our magnitude-based variable-length discretization procedure is organized around this calibration.

\paragraph{Unit calibration from dataset-level statistics.}
For each NCS we compute its total translation $L_{\text{trans}} = \sum_i \lVert \mathbf{p}_{i+1} - \mathbf{p}_i \rVert$ (integrated path length) and total rotation $L_{\text{rot}} = \sum_i \sqrt{\Delta\text{yaw}_i^2 + \Delta\text{pitch}_i^2}$ (in degrees). We then sweep candidate values of the translation unit $\Delta_{\text{trans}}$ and inspect, for each candidate, the induced distribution of sequence length $T$ across the entire corpus; the smallest unit for which $T$ lies in a training-friendly range is retained. The rotation unit $\Delta_{\text{rot}}$ has a natural absolute scale and is fixed throughout. Our final release adopts
\[
\Delta_{\text{trans}} = 0.05,\qquad \Delta_{\text{rot}} = 5^\circ.
\]

\paragraph{Mixed-motion accumulator and bin boundaries.}
With $\Delta_{\text{trans}}$ and $\Delta_{\text{rot}}$ fixed, we scan an NCS in temporal order and track a single scalar budget
\[
m_i = \frac{\lVert \mathbf{p}_{i+1} - \mathbf{p}_i \rVert}{\Delta_{\text{trans}}}
      \,+\,
      \frac{\sqrt{\Delta\text{yaw}_i^2 + \Delta\text{pitch}_i^2}}{\Delta_{\text{rot}}}.
\]
The budget is accumulated frame by frame and a new bin is closed whenever the running sum reaches $1.0$; any residual below $1.0$ at the end of the segment is merged into the last bin. This yields a time-synchronous partitioning of the NCS in which every bin corresponds to an approximately unit-magnitude motion, jointly measured in translation and rotation.

\paragraph{Per-bin W/A/S/D token.}
Within each bin we emit one translation token and one rotation token by evaluating the net displacement $(\Delta x, \Delta y)$ and the net rotation $(\Delta\text{yaw}, \Delta\text{pitch})$ accumulated inside the bin. For either stream, the principal axis is first identified by the larger absolute magnitude and, if the secondary axis exceeds the principal axis by a ratio of $0.3$ (translation) or $0.5$ (rotation), the two keys are combined into a diagonal token such as \texttt{WD}. To suppress jitter, we additionally reject a bin-level token as $\texttt{N}$ when (i) its total magnitude is below $10\%$ of the corresponding unit, or (ii) the net-to-accumulated ratio is below $0.4$, a condition characteristic of back-and-forth oscillation. A whole NCS whose total translation and rotation both stay below static thresholds ($0.02$ in pose units and $2.5^\circ$) collapses to a single \texttt{N} token on each stream.

\paragraph{Output.}
The procedure produces two equally long W/A/S/D sequences---one for translation and one for rotation---in which each token carries an approximately fixed physical meaning of ${\sim}0.05$ in translation or ${\sim}5^\circ$ in rotation. These two streams are stored together with their segment-level metadata (\textit{e.g.}, $L_{\text{trans}}$, $L_{\text{rot}}$, bin boundaries) alongside the raw annotation JSON and are used as the ground-truth action sequence $\mathbf{a}_{1:T}$ throughout this work.

\subsection{Instruction Annotation and Verification}
\label{app:annot}

This subsection details how the natural-language task instruction $\ell$ is produced for each NCS. Following the two-track design in Sec.~\ref{sec:dataset}.2 Step~3, we generate two complementary annotations per NCS---a \emph{Detailed Movement Instruction} and, whenever feasible, a \emph{High-Level Goal-Navigation Caption}---and adopt a higher-level-first, detailed-fallback policy when assembling the final training corpus. Both tracks are implemented via automatic prompting of the same frontier VLLM, Qwen3.5-27B~\citep{bai2025qwen3}, served with vLLM tensor parallelism.

\paragraph{Detailed Movement Instruction.}
For each NCS we uniformly sample frames from the corresponding video segment at $2$~fps and feed them, together with an overall direction descriptor derived from $\mathbf{a}_{1:T}$ (\textit{e.g.}, \texttt{"left-forward"}), into the VLLM under the prompt shown below. 
The VLLM is asked to emit a structured JSON record with four fields---\texttt{annotatable}, \texttt{task\_description}, \texttt{move\_target}, and \texttt{move\_type}---together with a self-reported \texttt{direction\_consist} flag indicating whether the described motion is consistent with the direction descriptor.

\begin{center}
\begin{tcolorbox}[colback=gray!00,
                  colframe=black,
                  arc=1.5mm, auto outer arc,
                  breakable,
                  left=0.9mm, right=0.9mm,
                  boxrule=0.9pt,
                  title = {Full prompt for annotating the Detailed Movement Instruction.},
                  label = {box:prompt-detailed}]
\scriptsize
\begin{verbatim}
Please **strictly** follow all instructions below.
Your output **must** be a valid JSON (or Python dictionary) object and **must not**
contain any additional explanations, comments, punctuation remarks, or code-block
wrappers.

---

## Annotation Task

You are given a **trajectory sub-segment** extracted from a longer video. This
sub-segment corresponds to a continuous time interval during which the primary
character maintains a **consistent movement direction**. The movement direction
metadata is provided alongside the video frames.

Your task is to annotate the **task instruction** executed by the **primary
character** within this sub-segment, along with the associated **movement target
information**.

We assume that the target's task intent is reflected by its overall movement
direction. Since the sub-segment already exhibits a unified movement direction,
you should produce exactly one task annotation for the entire sub-segment.

---

## Definition of Task Instructions

- A **task instruction** refers to a **high-level behavioral objective** that
  explains *why* the character is moving in the observed direction during this
  sub-segment.
  Examples include: *navigating toward a door*, *approaching an enemy*,
  *retreating to cover*, *exploring a corridor*, *moving toward a checkpoint*.

- Task instructions are identified by reasoning about:
  - What goal is the character attempting to achieve during this sub-segment?
  - What destination or object is the character moving toward (or away from)?
  - What high-level intent does the observed movement serve?

- **Distinction between Action and Task**:
  - **Action**: A short-term, atomic behavior (e.g., turning, jumping, opening
    a door).
  - **Task**: A higher-level objective that may involve multiple actions,
    defined by a consistent goal and a clear target.

---

## Key Annotation Principles

### Primary Character Priority
First, identify the primary character (from either a first-person or
third-person perspective) and center all annotations on this character's
behavior.
If multiple characters are present without a clear hierarchy, or if no character
appears in the entire video segment (e.g., the video only contains environmental
or scene information), the annotation should instead be based on the movement
of the main camera viewpoint.

### Visual Content Only
Task information must be derived **exclusively from the visual content of the
video frames**.
Ignore all on-screen text, UI elements, HUD overlays, subtitles, or displayed
task prompts. Even if the video explicitly shows textual instructions, you must
deliberately ignore them and rely solely on observable visual evidence.

### Precise and Unique Task Description
The `task_description` must be **precise enough to uniquely identify the
character's movement process** within this sub-segment. This is achieved
through two mechanisms:

1. **End-State Anchoring**: Constrain the description using information
   observable at the **final frame** of the sub-segment. The movement target
   should ideally be a specific, identifiable object or location that the
   character arrives at or approaches by the end.

2. **Directional Disambiguation**: When multiple similar objects exist in the
   scene, use directional cues (e.g., *the door on the left*, *the cabinet to
   the front-right*) to uniquely identify the target.

3. **Initial-Frame Visibility Requirement**: When referencing a specific object
   as the movement target, that object **must be visible in the first frame**
   of the sub-segment. If the target object only appears later in the
   sub-segment and is not visible at the start, it **must not** be used as the
   anchoring reference. In such cases, use a more general spatial description
   (e.g., *moving toward the end of the corridor*) or directional reference
   instead.

4. **Fallback**: If no unique target can be identified (e.g., the character is
   wandering or exploring without a clear destination), describe the movement
   in general terms and set `move_target` to `null`.

### [!] Direction Consistency in task_description (CRITICAL REQUIREMENT)
The `task_description` **must explicitly include directional words that match
every component of the provided `{{MOVEMENT_DIRECTION}}`**. This is a hard
requirement that cannot be omitted.

- The `{{MOVEMENT_DIRECTION}}` value is composed of one or more direction
  components joined by `-` (e.g., `"left-forward"`, `"right-backward-up"`).
- **Each component must appear explicitly in the `task_description`** using
  natural language directional expressions. The mapping is:
  - `forward`  -> must include words such as *forward*, *ahead*, *front*
  - `backward` -> must include words such as *backward*, *back*, *rear*,
                  *retreat*
  - `left`     -> must include the word *left*
  - `right`    -> must include the word *right*
  - `up`       -> must include words such as *up*, *upward*, *upper*, *above*,
                  *ascending*
  - `down`     -> must include words such as *down*, *downward*, *lower*,
                  *below*, *descending*

- **Examples**:
  - `{{MOVEMENT_DIRECTION}} = "forward"` -> `task_description` must explicitly
    contain *forward / ahead / front*
  - `{{MOVEMENT_DIRECTION}} = "left-forward"` -> `task_description` must
    explicitly contain both *left* and *forward / ahead / front*
  - `{{MOVEMENT_DIRECTION}} = "right-backward-up"` -> `task_description` must
    explicitly contain *right*, *backward / back / rear*, and *up / upward /
    ascending*

- This directional information must be woven **naturally into the sentence**
  (not appended mechanically). The description should still read as a fluent,
  high-level task instruction.

- [OK] Good example (`{{MOVEMENT_DIRECTION}} = "left-forward"`):
  *"The character advances toward the front-left, heading for a metal door at
   the end of the left corridor."*

- [X] Bad example (`{{MOVEMENT_DIRECTION}} = "left-forward"`, missing direction
  words):
  *"The character heads toward a metal door at the end of the corridor."*

### Diversity of Expression
Use **varied and natural language** to describe the task instruction. Avoid
repetitive sentence patterns across different annotations. For example, the
following expressions all describe similar movement but with different
phrasing:

- *The character advances through the corridor toward a wooden door at the far
   end.*
- *Navigating along a dimly lit hallway, the character heads for the exit
   ahead.*
- *The character proceeds down the passage, approaching the doorway visible in
   the distance.*
- *Moving through the narrow corridor, the character makes their way to the
   door at the front.*

Vary sentence structure, vocabulary, and the order of information (target
first vs. action first, etc.) to ensure annotation diversity.

### High-Level Semantics First
Focus on **what the character is trying to achieve**, not on low-level actions.
The `task_description` must be a **task-level abstraction**, not a simple
enumeration of actions.

### Uncertainty Handling
If the character's objective cannot be reliably inferred from the visual
content, set `"annotatable": false` and leave all other fields as `null`. Do
not speculate.
Additionally, when applying **End-State Anchoring** using a **movement target
location**, you must verify whether the referenced movement target visible in
the final frame also appears in the **first frame** of the sub-segment. If the
target location does **not** appear in the first frame, set
`"annotatable": false` and leave all other fields as `null`.

---

## Movement Target Information

### Movement Target (`move_target`)
- Identify the **specific spatial target** that the character moves toward (or
  remains at) during this sub-segment.
  - Describe the target relative to concrete objects or spatial references
    visible in the scene (e.g., *in front of the red door*, *beside the large
    tree on the left*).
  - The referenced object **must be visible in the first frame**. If it only
    appears mid-way through the sub-segment, do not use it as the anchor.
  - If the character remains stationary, describe the position where the
    character stays.
  - If no clear target exists (e.g., wandering, patrolling), output `null`.

### Movement Type (`move_type`)
- Summarize the character's **overall movement direction** during this
  sub-segment.
  - Use coarse-grained descriptions: *moving forward*, *moving to the
    front-left*, *turning right*, *remaining stationary*, *retreating
    backward*, etc.
  - If the movement involves mixed directions, summarize with the dominant
    direction (e.g., *moving toward the front-right*).

### Direction Consistency (`direction_consist`)
- Set to `true` if the `move_type` you annotated is **semantically consistent**
  with the provided `{{MOVEMENT_DIRECTION}}`; otherwise set to `false`.
- Set to `null` if the sample is not annotatable (`"annotatable": false`).
- This field serves as a self-check: if your annotated `move_type` conflicts
  with the provided `{{MOVEMENT_DIRECTION}}`, re-examine your annotation
  before finalizing.

---

## Examples

### Case 1
**Input `{{MOVEMENT_DIRECTION}}`**: `"left-forward"`
{
  "annotatable": true,
  "perspective": "Third-person",
  "task_description": "The character navigates through a nighttime campsite,
                       moving toward the front-left to reach the entrance of
                       a tent situated ahead and to the left.",
  "move_target": "the tent entrance to the front-left",
  "move_type": "moving toward the front-left",
  "direction_consist": true
}

### Case 2
**Input `{{MOVEMENT_DIRECTION}}`**: `"forward"`
{
  "annotatable": true,
  "perspective": "First-person",
  "task_description": "Advancing along a sunlit forest trail, the character
                       pushes forward toward a wooden bridge visible directly
                       ahead.",
  "move_target": "the wooden bridge directly ahead",
  "move_type": "moving forward",
  "direction_consist": true
}

### Case 3
**Input `{{MOVEMENT_DIRECTION}}`**: `"right-backward"`
{
  "annotatable": true,
  "perspective": "First-person",
  "task_description": "The character retreats backward and to the right from
                       an open battlefield, falling back toward a stone wall
                       located to the rear-right for cover.",
  "move_target": "the stone wall to the rear-right",
  "move_type": "moving toward the rear-right",
  "direction_consist": true
}

### Case 4
**Input `{{MOVEMENT_DIRECTION}}`**: `"right-forward-up"`
{
  "annotatable": true,
  "perspective": "First-person",
  "task_description": "The character climbs upward along a rocky slope, moving
                       forward and to the right toward a ledge visible at the
                       upper-right of the cliff face.",
  "move_target": "the ledge at the upper-right of the cliff face",
  "move_type": "moving toward the upper-right while ascending",
  "direction_consist": true
}

### Case 5
**Input `{{MOVEMENT_DIRECTION}}`**: `"forward"`
{
  "annotatable": true,
  "perspective": "First-person",
  "task_description": "The character roams forward through a dense forest area
                       without a clearly identifiable destination ahead.",
  "move_target": null,
  "move_type": "moving forward",
  "direction_consist": true
}

### Case 6 (not annotatable)
**Input `{{MOVEMENT_DIRECTION}}`**: `"left-forward"`
{
  "annotatable": false,
  "perspective": null,
  "task_description": null,
  "move_target": null,
  "move_type": null,
  "direction_consist": null
}

---

## Input

`{{VIDEO_FRAMES}}` - A sequence of video frames corresponding to one trajectory
sub-segment. The sub-segment has a consistent primary movement direction.

`{{MOVEMENT_DIRECTION}}` - The dominant movement direction of this sub-segment
(e.g., "forward", "right-forward", "left"). This is provided as auxiliary
reference; the annotation must still be grounded in the visual content. **The
`task_description` must explicitly include directional words corresponding to
every component of this value.**

---

## Output Format

Return a JSON dictionary with the following structure:

{
  "annotatable": <bool>,
  "perspective": "<First-person / Third-person / Other, or null if not
                   annotatable>",
  "task_description": "<high-level description of the character's task
                        instruction, explicitly containing directional words
                        matching all components of {{MOVEMENT_DIRECTION}},
                        or null>",
  "move_target": "<the character's movement target, or null>",
  "move_type": "<the character's overall movement direction, or null>",
  "direction_consist": "<true if move_type is consistent with
                         {{MOVEMENT_DIRECTION}}, false if not, null if not
                         annotatable>"
}
\end{verbatim}
\end{tcolorbox}
\end{center}

\paragraph{Post-hoc direction-consistency check.}
To catch annotations that silently contradict the underlying action sequence, every record produced above is re-evaluated by an independent call that parses both the direction descriptor and the free-text \texttt{move\_type} into a pair of axis-level labels (forward/backward and left/right) and compares them. Any NCS whose two labels disagree on either axis is removed from the Detailed Movement Instruction pool.

\paragraph{High-Level Goal-Navigation Caption.}
The higher-level variant specifies only the navigation destination without any directional cues. Because it requires the target to be uniquely identifiable within the NCS, we produce it through a four-step grounding-and-verification loop:
\begin{enumerate}
    \item strip direction words from the \texttt{move\_target} field of the Detailed Movement Instruction to obtain a candidate target phrase;
    \item ground the candidate phrase on the initial observation $o_0$ with an open-vocabulary detector, keeping only cases in which exactly one bounding box is returned;
    \item ground the same phrase on the final frame of the NCS, again keeping only single-bbox cases;
    \item perform an LLM-based same-entity verification between the two boxes.
\end{enumerate}
Only NCSs that pass all four checks are admitted to this track, and the VLLM is prompted to generate a short, direction-free goal-navigation caption conditioned on the verified target phrase (\textit{e.g.}, \emph{``Head to the line of trees.''} or \emph{``Navigate to the wooden gate structure.''}).

\paragraph{Training-data construction policy.}
When assembling the final training corpus, we adopt a higher-level-first, detailed-fallback policy: we use the Goal-Navigation Caption as $\ell$ whenever it is successfully produced, and otherwise fall back to the Detailed Movement Instruction. The NCS-level share of each track in the released dataset is reported in Fig.~\ref{fig:SA_scene_1} and the accompanying statistics.

\paragraph{What the segment endpoint certifies.}
Because every trajectory-level metric is anchored at the endpoint of the
ground-truth NCS, it is worth stating precisely what that endpoint does
and does not guarantee. The segmentation itself is triggered by a change
of principal motion direction (App.~\ref{app:ncs}) rather than by a
verified arrival, so an NCS endpoint is \emph{not}, on its own, a
certificate that the agent reached a target object. The
goal--endpoint correspondence is instead established by the annotation
procedure above: an NCS is admitted to the High-Level Goal-Navigation
track only if a target object is identifiable in its \emph{final} frame
and survives all four grounding-and-verification checks, so that for
this track the endpoint is, by construction, a position at which the
instructed target has been reached. For the Detailed Movement
Instruction track no such claim is made: the instruction describes the
movement that the segment carries out, and the endpoint is simply where
that movement ends. The metrics are defined to respect this
distinction---they score the \emph{displacement consistency} between the
predicted and ground-truth endpoints rather than physical contact with
an object---as discussed in App.~\ref{app:metric_semantics}.

\subsection{Additional Statistical Analyses}
\label{app:scene_app}

\begin{figure}[!h]
    \centering
    \includegraphics[width=\linewidth]{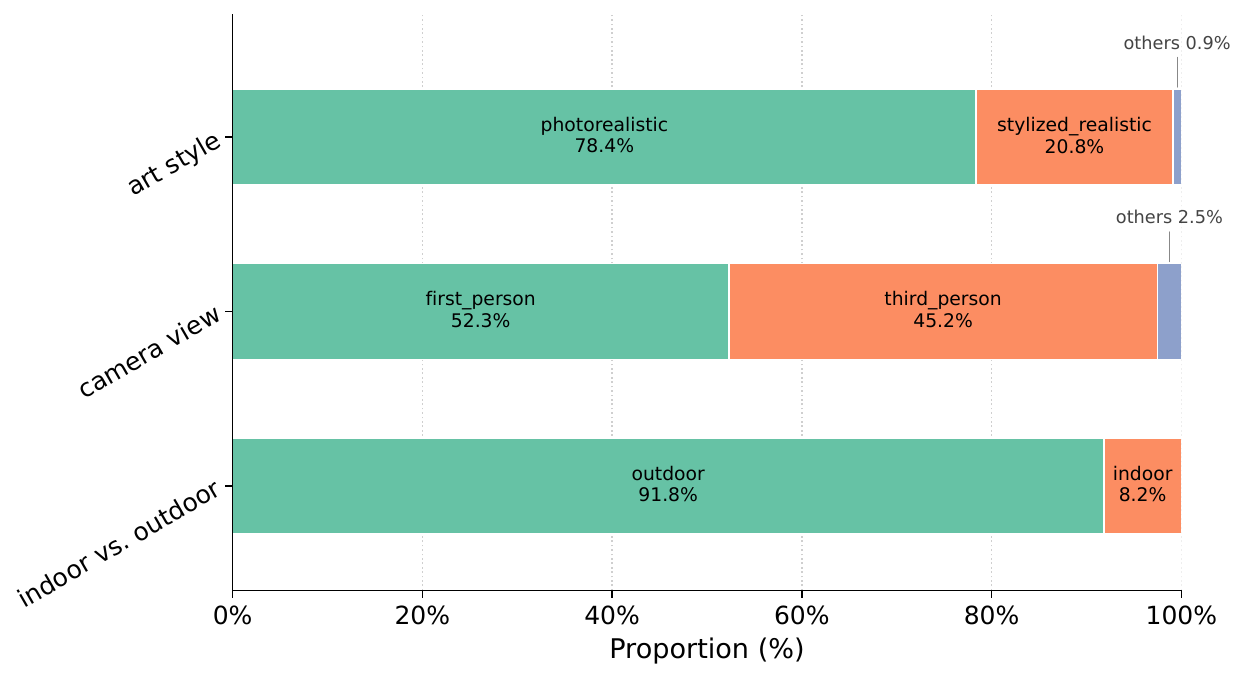}
    \caption{\textbf{Complementary statistical analysis of $o_0$ in EmbodiedWorld-200K}, along additional dimensions including camera viewpoint, art style, and other metadata axes.}
    \label{fig:SA_scene_app}
\end{figure}

This subsection complements Sec.~\ref{sec:dataset}.3, in which the diversity of the visual observation $o_0$ is analyzed through a keyword word cloud together with two coarse-grained breakdowns along the game-genre and scene-category axes (Fig.~\ref{fig:SA_scene_1}). To provide a more comprehensive view of environment diversity, we further report per-dimension distributions of $o_0$ along additional metadata axes, including \textbf{camera viewpoint}, \textbf{art style}, and other auxiliary attributes, as summarized in Fig.~\ref{fig:SA_scene_app}. These plots reinforce the conclusion drawn in the main text: EmbodiedWorld-200K spans a broad and well-balanced distribution of 3D environments along every examined axis, providing a solid visual foundation on which EWAs can acquire open-world priors through offline pretraining.

\section{More Details About Evaluation Metrics}
\label{app:metrics}

This appendix complements Sec.~\ref{sec:metric} and specifies the full algorithmic details of the four evaluation metrics. Following the organization in the main paper, we first describe the two \textbf{token-level} metrics --- \textsc{TM} and \textsc{DirAcc} --- that assess the predicted action sequence on a per-step basis, and then describe the two \textbf{trajectory-level} metrics --- \textsc{nDTW} and \textsc{NE} --- that evaluate the reconstructed 3-D trajectory as a whole. We also specify the shared pre-processing rules (length alignment and parse-failure fallback) that apply to all four metrics.

\paragraph{Shared pre-processing.}
For every evaluation sample we parse the predicted and ground-truth action sequences into translation and rotation token lists. Before computing any token-level metric, both sequences are right-padded with the no-op token \texttt{N} to a common length $L = \max(|\hat{\mathbf{a}}|, |\mathbf{a}|)$, so that length mismatches are penalized symmetrically. A sample whose prediction fails JSON parsing is replaced by a single-token no-op $[\texttt{N}]$ on each stream and scored through the identical pipeline, so that no samples are dropped from the evaluation pool.

\subsection{Token-level metrics}

\paragraph{TM (Token Match, $\uparrow$).}
TM measures how well the predicted keys at each time step match the ground-truth keys, and is robust to the fact that a combined token such as \texttt{WD} only partially overlaps with a simpler token such as \texttt{W}. Concretely, we treat each token as a subset of $\{\texttt{W}, \texttt{A}, \texttt{S}, \texttt{D}\}$, with $\texttt{N}$ identified with $\varnothing$. Let $g_t$ and $p_t$ denote the ground-truth and predicted token sets at step $t$; we compute a per-step set-level $F_1$,
\[
P_t = \frac{|g_t \cap p_t|}{|p_t|},\quad
R_t = \frac{|g_t \cap p_t|}{|g_t|},\quad
F_1^{(t)} = \frac{2\,P_t\,R_t}{P_t + R_t},
\]
with the convention that $F_1^{(t)} := 1$ when $g_t = p_t = \varnothing$ and $F_1^{(t)} := 0$ when exactly one side is empty. TM is obtained by averaging $F_1^{(t)}$ over all $L$ steps on each stream, and then averaging the translation and rotation scores.

\paragraph{DirAcc (Direction Accuracy, $\uparrow$).}
DirAcc probes whether the \emph{overall} direction of motion is captured correctly on four independent axes: translation forward/backward, translation left/right, rotation pitch, and rotation yaw. For each axis we derive a single categorical label in $\{\text{positive}, \text{negative}, \text{neutral}\}$ by comparing the counts of the two opposing keys across the full sequence (\textit{e.g.}, count of \texttt{W} versus count of \texttt{S} on the translation FB axis). DirAcc is the mean four-axis agreement between the predicted labels and the ground-truth labels over the evaluation set.

\subsection{Trajectory-level metrics}

\paragraph{Trajectory reconstruction.}
All trajectory-level metrics are computed on 3-D trajectories reconstructed from the discrete W/A/S/D sequences with the dataset-calibrated physical units of $\Delta_{\text{trans}} = 0.05$ and $\Delta_{\text{rot}} = 5^{\circ}$. Starting from the origin $(x, y, z, \text{yaw}, \text{pitch}) = \mathbf{0}$, at each step we first apply the rotation token to update the heading, and then apply the translation token in the body-frame forward direction; composite tokens (\textit{e.g.}, \texttt{WD}) contribute to both of their component axes. This yields a pose sequence of length $|\mathbf{a}|{+}1$ for both the prediction and the ground truth.

\paragraph{nDTW (normalized Dynamic Time Warping, $\uparrow$).}
nDTW assesses \emph{path shape} while absorbing moderate misalignments in pace between the two trajectories. Standard dynamic time warping on the reconstructed 3-D point sequences yields a cumulative distance $\mathrm{DTW}(\hat{\mathbf{a}}, \mathbf{a})$, which we normalize by the ground-truth length and a distance-tolerance coefficient $\tau$,
\[
\mathrm{nDTW}(\tau) \;=\; \exp\!\left(-\,\frac{\mathrm{DTW}(\hat{\mathbf{a}}, \mathbf{a})}{|\mathbf{a}|\cdot\tau}\right).
\]
We report $\mathrm{nDTW} = \mathbb{E}_{\tau \in \{0.5, 1.0, 2.0\}} \mathrm{nDTW}(\tau)$ to cover strict, moderate, and loose deviation tolerances.


\paragraph{NE (Navigation Error, $\downarrow$).}
NE directly reports the planar endpoint displacement $\mathrm{NE} = d_{xy}$ in translation units, where $d_{xy}$ denotes the planar distance between the predicted and the ground-truth endpoint. Complementary to the shape-based nDTW, it provides a threshold-free, purely endpoint-based readout of whether the agent reproduces the goal-directed displacement described by the instruction.

\subsection{What the Metrics Do and Do Not Measure}
\label{app:metric_semantics}

The two trajectory-level metrics are deliberately complementary, and
reading them together is what makes the evaluation well-posed. We make
their division of labour, and its two main consequences, explicit here.

\paragraph{Path shape versus goal displacement.}
nDTW is a \emph{path-shape} measure: it rewards following the reference
trajectory and therefore penalizes an alternative route that reaches the
same destination by a different path. NE is \emph{path-agnostic}: it
compares only the two endpoints, so any route that arrives at the same
displacement receives the same score. Because we report both, the two
orthogonal questions---``did the agent arrive where the instruction
asked?'' (NE) and ``did it follow the reference path?'' (nDTW)---are read
out independently, and neither can silently dominate the other. In
particular, nDTW's penalty on valid alternative paths does not drive any
conclusion in this paper: every headline comparison is accompanied by
NE, which does not share that penalty. The converse also holds, which is
why we do not report NE alone: an agent that reaches the right endpoint
through an erratic detour is distinguishable only through nDTW.

\paragraph{Displacement consistency, not physical contact.}
As noted in App.~\ref{app:annot}, the endpoint of an NCS marks the end of
a direction-coherent movement, and only for the High-Level
Goal-Navigation track is it additionally verified to be a position at
which an identifiable target has been reached. Our endpoint metric is
defined to match this weaker guarantee: NE measures whether the agent
reproduces the overall goal-directed \emph{displacement} that the
instruction describes, not whether it physically touches the target
object. It is therefore insensitive to the precise contact geometry at
the goal, and robust to the residual localization error of the
monocular pose reconstruction from which the ground truth is derived.

\paragraph{Capabilities outside the scope of these metrics.}
All four metrics score an instruction-conditioned action sequence
against a reference sequence. They consequently do \emph{not} evaluate
object interaction or manipulation, recovery from execution errors,
long-horizon task decomposition, or any behaviour that changes the state
of the environment; nor do they capture exploration under partial
observability, since the instruction and the initial observation are
given. We regard these as complementary axes left to future benchmarks
(App.~\ref{app:limitations}), and the numbers reported here should be
read as evidence about open-world navigation-planning specifically.

\section{More Details About Stage~I: Pretraining on EmbodiedWorld-200K}
\label{app:stage1}

This appendix complements Sec.~\ref{sec:baselines} and Sec.~\ref{sec:eval}.1 by providing the full implementation details of Stage~I.

\paragraph{Training recipe and compute.}
We instantiate the baseline with \texttt{Qwen3-VL-4B-Instruct}~\citep{bai2025qwen3} and perform LoRA~\citep{hu2022lora} fine-tuning on the training split of EmbodiedWorld-200K. The LoRA adapters are attached to all linear layers of the backbone with rank $r=32$, scaling factor $\alpha=64$, and dropout $0.005$. Each training sample pairs an initial observation $o_0$ with the instruction $\ell$ and the ground-truth action sequence $\mathbf{a}_{1:T}$; the image pixel budget is capped between $12{,}544$ and $451{,}584$ pixels and the sequence length is capped at $4{,}096$ tokens. We use AdamW with learning rate $1\mathrm{e}{-4}$, a cosine schedule with $5\%$ linear warm-up, weight decay $0.01$, and bf16 mixed precision with DeepSpeed ZeRO-3. The effective batch size is $256$ (per-device batch size $8\times$ gradient-accumulation steps $4\times 8$ GPUs), and we train for two epochs. The full Stage~I run is carried out on $8\times$ NVIDIA H20 GPUs with a wall-clock time of approximately $5$ hours ($\sim\!40$ GPU-hours in total).

\paragraph{Baseline system prompt.}
\label{app:prompt}
At inference time, the baseline receives $(o_0, \ell)$ through a fixed system prompt that specifies the task, the discrete W/A/S/D action space, the physical meaning of each unit token, and the required JSON output format. The same prompt is used throughout training and evaluation, and is reproduced verbatim below.

\begin{center}
\begin{tcolorbox}[colback=gray!00,
                  colframe=black,
                  arc=1.5mm, auto outer arc,
                  breakable,
                  left=0.9mm, right=0.9mm,
                  boxrule=0.9pt,
                  title = {Baseline system prompt used by the Stage-I model for open-world embodied planning.},
                  label = {box:stage1-prompt}]
\scriptsize
\begin{verbatim}
You are a game-playing agent. Your task is to navigate the environment based
on visual observations and user instructions.
Task: Based on the given image and the goal-navigation instruction, predict
the discrete movement operation sequence that best fulfills the instruction.
Movement Instruction: "{{INSTRUCTION}}".
Action Space: for translation, 'W' (forward), 'A' (left), 'S' (backward),
'D' (right), or 'N' (no action); for camera rotation, 'W' (up), 'A' (left),
'S' (down), 'D' (right), or 'N' (no action). You can combine keys for
diagonal movement (e.g., 'WA' for forward-left).
Note: the output sequence length is NOT fixed. Each action token represents
one unit of physical motion (~0.05 length for translation, ~5 degrees for
rotation). Typical length is around 12, but can range from 1 to 40 depending
on the scale of the movement. 'translation' and 'rotation' must have the
same length.
Before writing out the action sequences, first decide how many action steps
are needed based on the spatial distance to the goal and the scale of the
described movement. Put this number as the FIRST field "action_num", then
emit the two lists whose lengths MUST strictly equal "action_num".
Output format: strictly a JSON object, no other text:
{
  "action_num": <int>,
  "translation": ["key1", "key2", ...],
  "rotation":    ["key1", "key2", ...]
}
\end{verbatim}
\end{tcolorbox}
\end{center}

\subsection{Effect of the Stage-I Pretraining Data Scale}
  \label{app:stage1_scale}

  To understand how much of EmbodiedWorld-200K is actually needed to
  drive the Stage-I gains reported in Sec.~\ref{sec:eval}.1, we
  additionally ablate the training-set size by pretraining the baseline
  on $\{1/3, 2/3, 3/3\}$ uniformly sub-sampled shards of the full
  training split while keeping the remaining training recipe and the
  held-out evaluation split exactly the same as in                                                        
  App.~\ref{app:stage1}.
                                                                                 
  As shown in Tab.~\ref{tab:sft_data_scale}, every metric improves
  substantially over the zero-shot \emph{Base} model at every data
  scale, confirming that the
  Stage-I gains are driven by the scale of EmbodiedWorld-200K rather
  than by any specific subset. A closer look at the per-metric gains
  reveals two distinct regimes: the trajectory-level metrics (nDTW and
  NE) already capture most of their improvement with only
  one-third of the corpus (nDTW $67.1\!\to\!83.0$,
  NE $51.7\!\to\!40.1$) and then plateau, while TM
  keeps growing steadily across all data scales
  ($54.9\!\to\!56.3\!\to\!59.7\!\to\!62.0$); DirAcc, which only probes
  the overall direction on each axis, is largely saturated already at
  the backbone level and stays flat throughout. In other words, the
  macroscopic navigation structure of open-world 3D scenes---when to
  turn, how far to travel, and where to arrive---is relatively
  data-efficient on top of a strong VLLM prior, whereas fine-grained
  per-step precision continues to benefit from additional data; both
  observations are consistent with the view that EmbodiedWorld-200K
  provides the scale and diversity needed to elicit open-world embodied
  navigation-planning priors from a frontier VLLM.

  \begin{table}[h]
  \centering
  \small
  \caption{Performance on the EmbodiedWorld-200K evaluation split,
  before (\emph{Base}) and after Stage-I pretraining with $\{1/3, 2/3,
  3/3\}$ of the full training corpus.}
  \label{tab:sft_data_scale}
  \begin{tabular}{lcccc}
  \toprule
  \textbf{Setting} & \textbf{TM $\uparrow$} & \textbf{DirAcc $\uparrow$} & \textbf{nDTW $\uparrow$} &
  \textbf{NE $\downarrow$} \\
  \midrule
  Base                & 54.9 & 70.1 & 67.1 & 51.7 \\
  +SFT ($1/3$ data)   & 56.3 & 70.1 & 83.0 & 40.1 \\
  +SFT ($2/3$ data)   & 59.7 & 71.8 & 84.4 & 36.7 \\
  +SFT ($3/3$ data)   & 62.0 & 72.0 & 83.6 & 35.6 \\
  \bottomrule
  \end{tabular}
  \end{table}

\section{More Details About Stage~II: World-Model-Based RL Post-Training}
\label{app:stage2}

This appendix complements Sec.~\ref{sec:baselines} and Sec.~\ref{sec:eval}.2 and provides the full implementation details of Stage~II. As stated in the main text, Stage~II treats a frontier video world model (LingBot-World-Fast~\citep{team2026advancing}) as a low-cost proxy for the target environment (GTA5), and post-trains the Stage-I actor with DPO~\citep{rafailov2023dpo} on preference pairs collected via closed-loop rollouts. Turning this conceptually simple recipe into a reproducible pipeline, however, requires careful design at every step. We organise the implementation into four tightly coupled stages---(i) initial observation and candidate action collection in the real game environment, (ii) rollout inside the world model, (iii) VQA-based reward scoring and preference-pair construction, and (iv) DPO optimisation---and describe each below.

\paragraph{Step 1: Initial observation and candidate action collection.}
We use the real GTA5 environment solely to (a) harvest diverse initial observations $o_0$ and (b) derive, for each such $o_0$, a candidate action sequence that will drive the subsequent world-model rollout. To obtain a broad distribution of starting states without costly full episode resets, we adopt a \textbf{backward-reset randomisation} strategy: at the beginning of each trajectory, we roll out the actor for a short reverse-navigation horizon in the game, interleaved with stochastic yaw perturbations, so that every new trajectory starts from a fresh and semantically varied location.

Given $o_0$, the corresponding navigation task is not manually specified; instead, we prompt a VLLM (Qwen3.5-27B) with a dedicated two-step instruction-generation prompt---reproduced verbatim below---which first identifies reachable nearby objects in $o_0$ and then produces a concise, direction-free goal-navigation instruction $\ell$ anchored on one of them. This adaptive instruction-generation scheme ensures that $\ell$ stays semantically aligned with the current scene, and that the generated instruction carries the same style as the Goal-Navigation Captions used during Stage~I pretraining. The actor then maps $(o_0, \ell)$ to a candidate action sequence; to broaden the exploration space at rollout time, we use stochastic decoding during actor inference (temperature sampling with $T{=}0.3$ in our default setting), so that the same $(o_0, \ell)$ can yield mutually distinct candidate sequences. We deliberately use the Stage-I SFT actor, rather than the raw base VLLM, at this step: as noted in Sec.~\ref{sec:eval}.2 of the main text, the SFT actor already possesses a basic embodied-planning ability, so that its sampled sequences contain a non-trivial fraction of correct decisions and---combined with the randomised decoding above---yield rollouts of sufficient quality to form informative preference pairs; starting from the base model would produce rollouts too noisy to support effective preference learning, underscoring the necessity of Stage~I. Each candidate action sequence is finally executed in GTA5, and the full per-step screenshots together with the action trace are recorded as the reference input to Step~2.

\begin{center}
\begin{tcolorbox}[colback=gray!00,
                  colframe=black,
                  arc=1.5mm, auto outer arc,
                  breakable,
                  left=0.9mm, right=0.9mm,
                  boxrule=0.9pt,
                  title = {Two-step prompt used to adaptively generate the navigation task instruction $\ell$ from the initial observation $o_0$.},
                  label = {box:stage2-instr-prompt}]
\scriptsize
\textbf{Stage A --- Nearby-object identification.}
\begin{verbatim}
You are analyzing a screenshot from a 3D open-world game (similar to GTA5).

Look at the image carefully and list 5-8 distinct objects, landmarks, or
locations that are NEARBY or at MEDIUM distance in the scene. The agent
needs to be able to REACH these objects within about 12 walking steps
(~12 seconds).

DISTANCE RULES (very important):
- ONLY include objects that are CLOSE or at MEDIUM distance (roughly within
  10-30 meters in game world).
- PREFER objects in the near-to-mid foreground of the image (occupying a
  noticeable portion of the frame).
- DO NOT include distant skyline objects, far-away mountains, or tiny
  objects on the horizon.
- DO NOT include the sky, clouds, or background scenery that cannot be
  physically reached.
- DO NOT include objects that the player character is already standing on
  or touching.

Good examples: a nearby car, a door 10 meters ahead, a staircase to the
side, a street lamp, a dumpster, a fence, a parked vehicle, a shop
entrance.
Bad examples: a distant skyscraper, faraway hills, the sun, a building
barely visible on the horizon.

Focus on:
- Nearby buildings, structures, vehicles, signs
- Street furniture (benches, lampposts, trash cans, mailboxes)
- Interactive objects (doors, stairs, ramps) within walking distance
- Vehicles parked nearby

Output ONLY a valid JSON object:
{
  "objects": [
    {"name": "short description of object",
     "visual_features": "brief appearance details",
     "estimated_distance": "near/medium"},
    ...
  ]
}
\end{verbatim}

\vspace{0.4em}
\textbf{Stage B --- Goal-navigation instruction generation.}
\begin{verbatim}
You are writing a goal-navigation instruction for a game-playing AI agent.

The agent is viewing the current game scene. Based on the selected NEARBY
target object below, write a concise navigation instruction that tells the
agent WHERE to go. The target should be reachable within about 12 walking
steps.

Target object: "{{TARGET_NAME}}" ({{TARGET_FEATURES}})

CRITICAL RULES:
1. Do NOT use any directional words: forward, backward, left, right,
   ahead, behind, front, back, up, down.
2. Do NOT describe the path or route.
3. ONLY describe the destination target. You may include visual features
   of the target.
4. Use varied phrasing: "Go to...", "Navigate to...", "Move toward...",
   "Reach...", "Head to...", "Approach...".
5. Single sentence only.
6. The target must be a NEARBY, physically reachable object - not a
   distant landmark.

Output ONLY a valid JSON object:
{
  "instruction": "your navigation instruction here",
  "target_object": "{{TARGET_NAME}}"
}
\end{verbatim}
\end{tcolorbox}
\end{center}

\paragraph{Step 2: Rollout inside the video world model.}
Each $(o_0, \mathbf{a}_{1:T})$ tuple from Step~1 is repackaged into the input format expected by LingBot-World-Fast~\citep{team2026advancing}, consisting of three components: the raw image $o_0$, the actor-planned action sequence $\mathbf{a}_{1:T}$, and a free-form scene-caption prompt---together forming the input to the world model. A subtle but important detail is that, when feeding $\mathbf{a}_{1:T}$ to the world model, we further calibrate the execution duration of every individual action so that the displacement/rotation magnitude it produces inside the world model matches the physical magnitude of a unit action in the real GTA5 environment; only under such a magnitude alignment can the world model faithfully serve as a behavioural proxy for GTA5. The world model is then run in batch mode at a fixed $480\!\times\!832$ resolution with $13$ generated frames per unit action. For every executed step, we additionally save one representative frame from the generated video as the post-step state observation, which will in turn be consumed by the VQA-based reward scoring in Step~3.

\paragraph{Step 3: VQA-based reward and preference-pair construction.}
After rollouts are produced, we cast reward modelling as visual question answering~\citep{rocamonde2024vlmrm,wang2024rlvlmf}. For each rollout, we prompt Qwen3.5-27B (served via vLLM with tensor parallelism) to (i) ground the target object referenced by $\ell$ on the initial frame $o_0$, (ii) overlay the predicted target bounding box as a visual anchor, and (iii) at every step $t\!\in\![1,T]$ of the rollout, answer a structured question that evaluates whether the agent is approaching the target in the post-step state frame (saved in Step~2). The two prompts used in this procedure---one for target grounding and one for QA-based scoring---are reproduced verbatim in Tab.~\ref{box:stage2-reward-prompt}. Each scored step returns a tuple $(\text{score}, \text{confidence}, \text{parse\_ok})$, and is filtered on parse validity. With the per-step QA scores in hand, we can readily construct preference-pair samples for DPO training~\citep{hu2026blameerrantwords}.

\begin{center}
\begin{tcolorbox}[colback=gray!00,
                  colframe=black,
                  arc=1.5mm, auto outer arc,
                  breakable,
                  left=0.9mm, right=0.9mm,
                  boxrule=0.9pt,
                  title = {Prompts used for VQA-based reward scoring in Step~3: (A) target grounding on $o_0$, and (B) per-step scoring of the post-step state frame.},
                  label = {box:stage2-reward-prompt}]
\scriptsize
\textbf{Stage A --- Target grounding on the initial frame.}
\begin{verbatim}
Locate the following object in this image: "{{TARGET_OBJECT}}"

Report the bounding box coordinates in JSON format like this:
[{"bbox_2d": [x1, y1, x2, y2], "label": "{{TARGET_OBJECT}}"}]

Rules:
- Coordinates MUST be in the range 0-1000 (relative permille, not pixels).
- If the object is NOT visible in the image, output an empty list: [].
- If multiple instances exist, report only the MOST PROMINENT / CLOSEST
  one to the camera (i.e., the largest one in the current view).
- Be precise - the bounding box should tightly enclose the target object.

Output ONLY a valid JSON list, no other text.
\end{verbatim}

\vspace{0.4em}
\textbf{Stage B --- Per-step VQA scoring.}
\begin{verbatim}
You are evaluating a game agent's navigation attempt in a GTA5-style
scene rendered in FIRST-PERSON view. The camera IS the player's
viewpoint; there is no visible character body.

The agent's task: Navigate to the target object "{{TARGET_OBJECT}}".

TARGET ANCHORING (important):
- In Image 1 (BEFORE), the target object is enclosed in a RED bounding
  box. That specific instance inside the red box is THE target - not
  any other visually similar object nearby.
- Image 2 (AFTER) is UNANNOTATED. Because the camera has moved during
  the action steps, the same target may appear at a different screen
  position, a different scale, partially occluded, or viewed from a
  different angle. Its lighting and visible surfaces may differ.
- Re-identify the target in Image 2 using surrounding context (nearby
  landmarks, relative layout, distinctive shape/texture). Do not switch
  to a different instance of the same class just because it happens to
  be closer or more centered.
- If after careful inspection you cannot locate the same target in
  Image 2, treat it as "lost".

You are given TWO images and ONE action summary:
- Image 1 (BEFORE, RED BOX = target): the first-person view before acting.
- Image 2 (AFTER):  the first-person view after executing the first
  t action steps of the rollout.
- MAIN ACTION DIRECTION: the aggregated camera-frame translation was
  "{{MAIN_DIRECTION}}".
  (options: forward / forward-left / forward-right / left / right /
   backward-left / backward / backward-right / stationary)

Evaluate the attempt on the following dimensions.
OUTPUT STRICT JSON ONLY - no extra text, no markdown fences.

{
  "target_visible_after":
      "clearly_visible" | "partially_occluded" | "barely_visible" | "lost",
  "size_change":
      "much_larger" | "larger" | "same" | "smaller" | "much_smaller",
  "relative_closer":
      "much_closer" | "slightly_closer" | "same"
      | "slightly_farther" | "much_farther",
  "direction_alignment":
      "aligned" | "partially_aligned" | "orthogonal" | "opposite",
  "arrival_state":
      "far" | "approaching" | "near" | "at_interaction_range" | "overshot",
  "scene_progressed":
      "significant" | "moderate" | "minor" | "stuck",
  "obstacle_ahead":
      "none" | "distant_wall" | "close_wall" | "colliding",
  "reasoning":
      "<one short sentence describing the key visual evidence>"
}

Judgment guidelines (FIRST-PERSON view):
1. The camera IS the player. Image 1 and Image 2 are first-person
   screenshots with no visible player body. Judge motion by how the
   scene (target + its surroundings) changes in frame.
2. "relative_closer" is a TEMPORAL comparison between Image 1 and Image
   2. Use target apparent size, screen coverage, perspective convergence,
   and how much foreground detail is visible.
3. "size_change" refers to the target object's pixel size in Image 2
   vs Image 1.
4. "direction_alignment" compares the MAIN ACTION DIRECTION against the
   direction from the image center of Image 1 to the target's screen
   position. Forward motion aligned with a target near the image
   center = "aligned". Backward motion while the target is ahead =
   "opposite".
5. "arrival_state" in first-person view:
     - "at_interaction_range": target fills a large portion of the
       screen or nearly touches the viewport edges.
     - "near": target is clearly the dominant object in frame but not
       yet touching the viewport edges.
     - "approaching": target has grown in size but is still mid-distance.
     - "far": target is small or distant in Image 2.
     - "overshot": the camera appears to have moved past the target.
6. "scene_progressed" judges OVERALL scene change across the two
   frames, not target-specific. Under stuck (e.g., walking into a
   wall) this should be "stuck".
7. "obstacle_ahead" describes what the camera sees DIRECTLY IN FRONT
   OF IT in Image 2 ("none" / "distant_wall" / "close_wall" /
   "colliding").
8. If the target cannot be reliably identified in either image, set
   "target_visible_after" = "lost" and other fields based on best
   guess.

Additional guidance for the ANNOTATED setting:
9.  Base your identification of the target in Image 2 on contextual
    cues rather than on the red box alone (the red box exists only in
    Image 1).
10. If the target fills most of Image 2 or touches the viewport edges,
    prefer "arrival_state" = "at_interaction_range" or "near".
\end{verbatim}
\end{tcolorbox}
\end{center}

\paragraph{Step 4: DPO optimisation.}
We post-train the Stage-I actor with DPO using LoRA adapters on the same Qwen3-VL-4B-Instruct backbone ($r=32$). Training uses the sigmoid DPO loss with $\beta=0.1$ and a frozen reference model, AdamW with learning rate $5\mathrm{e}{-6}$, a cosine schedule with $10\%$ warm-up, an effective batch size of $64$ (per-device $1\times$ gradient-accumulation $8\times$ $8$ GPUs), and a sequence length of $2{,}048$. We train for $2$ epochs in bf16 with DeepSpeed ZeRO-2 on $8\times$ NVIDIA H20 GPUs.

\section{Limitations}
\label{app:limitations}

While EmbodiedWorld-200K and the accompanying two-stage training
recipe provide a reusable foundation for studying Embodied World
Agents, our work still has two limitations.

\paragraph{3D virtual games as proxies for the physical world.}                                                                                               
As discussed in Sec.~\ref{sec:dataset}.1, we intentionally build EmbodiedWorld-200K from modern 3D open-world games, which provide visually rich environments, physically plausible dynamics, and diverse embodied behaviors under a unified interaction framework. They also constitute one of the few scalable sources of large-scale 3D interactive data accessible at academic cost.
Although the resulting corpus spans diverse scenes, viewpoints, art styles, and interaction patterns (Fig.~\ref{fig:SA_scene_1} and Fig.~\ref{fig:SA_scene_app}), these environments remain simulated, designer-authored worlds rather than direct observations of reality. Consequently, factors critical for real-world deployment---such as fine-grained physical interaction, sensor noise, uncontrolled lighting, and long-tail human--object interactions---are only partially captured.
Nevertheless, we believe this gap does not diminish the value of EmbodiedWorld-200K as a pretraining resource. Advances in modern simulation engines, including high-fidelity physics and rendering, are steadily narrowing the discrepancy between virtual and real environments. More importantly, many priors learned from EmbodiedWorld-200K---such as scene structure, goal-oriented navigation, and embodied exploration strategies---are largely domain-agnostic. In practice, EmbodiedWorld-200K can serve as a scalable open-world pretraining corpus, followed by lightweight adaptation to downstream physical platforms through real-world finetuning or interactive post-training.

\paragraph{Limited action granularity.}
To unify trajectories across more than 150 heterogeneous games, EmbodiedWorld-200K adopts a shared coarse action space consisting of W/A/S/D movement, camera rotation, and a no-op action (\texttt{N}). While sufficient for the embodied navigation and planning tasks studied in this work, this abstraction omits fine-grained interaction primitives such as object manipulation, tool use, UI interaction, combat, and other game-specific controls.
As a result, EWAs trained on EmbodiedWorld-200K primarily acquire capabilities in perception, reasoning, and navigation across diverse 3D environments, but do not yet support complex task compositions requiring dexterous manipulation or richer multimodal control.


\section{Broader Impact}                                                                                
\label{app:broader_impact}  

 EmbodiedWorld-200K, together with the accompanying two-stage training
  recipe, is primarily an infrastructure contribution for studying
  general-purpose embodied agents in open-world 3D environments. In line
  with NeurIPS guidance, we reflect on both the positive and the
  potentially negative societal impact that such a release may bring.

  \paragraph{Positive impact.}
  Open-world embodied-agent research has, until now, been fragmented
  across isolated robot platforms, proprietary simulators, and closed                                     
  gameplay pipelines, which significantly raises the entry barrier for
  academic and non-industrial researchers. By releasing a large-scale                                     
  dataset of over 200K annotated navigation-planning trajectories that
  spans 150+ 3D open-world games, together with a unified evaluation                                      
  split, reproducible baselines, and a concrete two-stage training
  recipe, we aim to lower this barrier and enable groups without access
  to industrial-scale data or robot fleets to participate in the study
  of Embodied World Agents. We also expect the scene diversity, the
  multi-level annotations, and the domain-invariant open-world priors
  distilled into our baseline to benefit adjacent areas that rely on
  general-purpose embodied understanding, including assistive robotics,
  interactive tutoring systems that ground language in visual 3D worlds,
  and accessible game AI for single-player experiences.

  \paragraph{Potentially negative impact and mitigations.}
  Because the released baseline is explicitly trained to execute
  open-world navigation instructions, it could in principle be misused
  as a component for automated manipulation in online games
  (\textit{e.g.}, bot-driven farming or cheating in multiplayer titles).
  To reduce this risk, the released model is trained only on the
  unified W/A/S/D abstraction rather than on platform-specific low-level
  controls, is not bundled with any game-specific input-injection
  interface, and is distributed under a research-use license (see
  App.~\ref{app:annot} for the annotation protocol and the project page
  for the license terms). All underlying assets---including the
  gameplay video source OGameData~\citep{che2024gamegen}, the pose
  estimator VIPE~\citep{huang2025vipe}, and the VLLMs employed for
  annotation and reward modelling---are credited and used within their
  original license terms, and no personally identifiable information or
  human-subject data is involved.


\newpage

\end{document}